\documentclass[final]{article}

\usepackage[english]{babel}

\usepackage[letterpaper,top=2cm,bottom=2cm,left=3cm,right=3cm,marginparwidth=1.75cm]{geometry}

\usepackage{amsmath}
\usepackage{dsfont}
\usepackage{graphicx}
\usepackage[colorlinks=true, allcolors=blue]{hyperref}
\usepackage{subcaption}
\usepackage[colorinlistoftodos,obeyFinal]{todonotes}
\usepackage{pifont}
\usepackage{wrapfig,lipsum,booktabs}
\usepackage{colortbl}

\newcommand{\cmark}{\ding{51}}%
\newcommand{\xmark}{\ding{55}}%

\title{System Resilience through Health Monitoring and Reconfiguration}
\author{Ion Matei, Wiktor Piotrowski, Alexandre Perez, Johan de Kleer, Jorge Tierno,\\ Wendy Mungovan, Vance Turnewitsch}

\begin{document}
\maketitle

\begin{abstract}\footnote{This material is based upon work supported by the Defense Advanced Research Projects Agecny (DARPA) under Contract No. HR001120C0179.}
We demonstrate an end-to-end framework to improve the resilience of man-made systems to unforeseen events. The framework is based on a physics-based digital twin model and three modules tasked with real-time fault diagnosis, prognostics and reconfiguration. The fault diagnosis module uses model-based diagnosis algorithms to detect and isolate faults and generates interventions in the system to disambiguate uncertain diagnosis solutions. We scale up the fault diagnosis algorithm to the required real-time performance through the use of parallelization and surrogate models of the physics-based digital twin. The prognostics module tracks the fault progressions and trains the online degradation models to compute remaining useful life of system components. In addition, we use the degradation models to assess the impact of the fault progression on the operational requirements. The reconfiguration module uses PDDL-based planning endowed with semantic attachments to adjust the system controls so that the fault impact on the system operation is minimized. We define a resilience metric and
use the example of a fuel system model to demonstrate how the metric improves with our framework.
\end{abstract}

\section{Introduction}
We describe an approach to system design that
dramatically improves the resiliency of the resultant system.  We illustrate this
approach with the task of designing resilient systems for future generations of autonomous naval ships.
Most physical systems in use today were designed with three
limiting assumptions: (1) the system designer had a severely
restricted set of conditions in mind from which the system could
recover from (e.g., feedback control systems to maintain stability or
direction), (2) human operators are expected to quickly
respond to unexpected events, and (3) maintenance technicians are available to
repair or modify the system as needed.  These limitations made sense
given the technologies available to designers at the time.  The result
of these limitations are designs with limited resilience: lack of ability
for the system to adapt in sufficient time to events completely
unexpected by the original designer.

We now live in a world quite different than the one in which most of
our systems were designed.  Computation has become
exponentially cheaper making it possible to embed extremely
sophisticated computation on board.  There has been an explosion in
the types of sensors readily available along with steep price
declines.  Artificial Intelligence (AI) and (Prognostics and Health Management) PHM algorithms have dramatically increased in
capability to the point where one can embed AI systems to perform
tasks usually performed by human personnel.  To leverage
these advances, the physical components of systems often need to be
adapted.  First, the physical systems must become sensor rich.
Second, the physical design must include additional reconfigurable
redundancy that the AI can modify, as needed.
Third, the physical structures supporting humans can greatly be
reduced because fewer are needed (less accommodation, food, access, etc.) In our approach, the AI embedded in the system is fully model-based.
We have designed a system creator: given a model of a system and the
environment it is to operate within our system creator automatically
creates the AI to be embedded in the physical system.  The combined
system will be able to respond far more quickly than previously to a
vastly larger set of unexpected events.

Given the system is now controlled by an AI, an interesting
question surrounds how to design the physical system differently to
take advantage of the new computational capabilities.  Some key
questions are: (1) where to introduce (or remove) sensors and (2)
where to add (or reduce) redundancy.  Physical redesign is outside of
the scope of this paper, however, the system described in this paper
is able to provide a metric to evaluate the resiliency of
a system.  Given the model of the physical system, and given we can
construct the AI automatically, we can simulate the combination on any
collection of events we choose.  The collective score on that
collection is a metric we can now maximize in a Design Space
Exploration approach \cite{10.5555/2023011.2023014}.

\subsection{Framework}
\begin{figure}[htp!]
  \centering
  \includegraphics[width=0.75\linewidth]{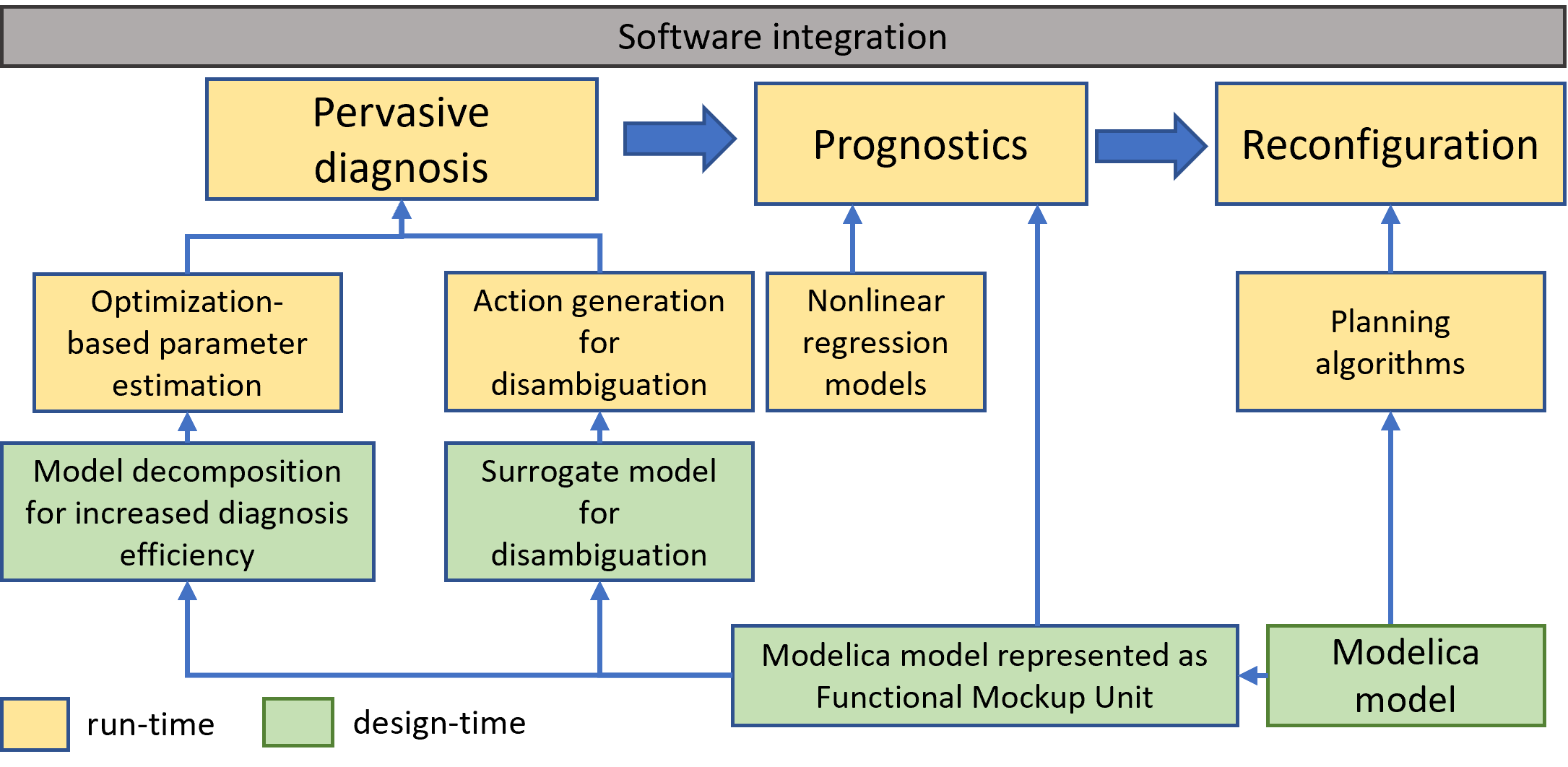}
  \caption{Block diagram of the automatically created embedded AI to ensure resilience.}
  \label{fig:approach diagram}
\end{figure}

System resilience is particularly important in remote applications, such as autonomous systems, where human experts are unavailable. Autonomous systems executing a remote mission require the capability to sense discrepancies in their behavior, reason about their causes and adapt to them to reduce the impact on their mission. We focus on system faults as the cause of observed discrepancies from the nominal behavior. We present an \textit{end-to-end} framework that uses health monitoring and reconfiguration to detect, reason and adapt to system faults, to ensure system resilience. The block diagram of the framework is shown in Figure \ref{fig:approach diagram}.
In our conception of a resilient system, algorithms supporting
diagnosis, prognosis and reconfiguration must continually operate as
the physical system such as a ship goes about its mission.  To
maximize resiliency we need those algorithms to be as general as
possible and not be tied to the specifics of the ship itself. We want
to avoid the fatal flaw of handcrafted resilience where the tacit
assumptions of the designer limit resiliency.  In our framework, at \emph{design-time}, we are
given only the model of the system expressed in a language like
Modelica.  Our software then constructs all the data structures needed
by the run time algorithms to achieve resilience: a functional mockup unit (FMU) \cite{blochwitz2011functional} to encode the Modelica model,  surrogate models
and model decompositions to improve efficiency, and PDDL models for reconfiguration  \cite{journals/corr/abs-1106-4561}.

At \emph{run-time}, the system resilience is ensured by three key modules responsible for health monitoring and reconfiguration. Health monitoring is implemented through the pervasive diagnosis and prognostics modules.  The reconfiguration module is responsible for adapting the system in response to fault effects. The pervasive diagnosis module uses sensor measurements and a model of the system to detect and isolate faults. We use a model-based diagnosis (MBD) approach, where a model of the system is coupled with an optimization algorithm to reason about system faults. The optimization algorithm estimates system parameters and compares them against their nominal values. An added feature of the pervasive diagnosis module is the decomposition of the system model  into subsystems that are diagnosed independently. The value of such a feature is improved scalability and the distributivity of the computational resources. A significant challenge in automated diagnosis is that for a set of off-nominal observations, the diagnosis solution can have high uncertainty, i.e., there are a large number of possible diagnoses. We minimize diagnosis uncertainty by designing limited interventions (i.e., sequence of control inputs) that drive the system through a sequence of states that makes the true diagnosis more evident.  We ensure scalability to real-time of the diagnosis process by (i) replacing the physics-based model with a surrogate model endowed with automatic-differentiation constructs. The prognostics module enables the prediction of fault evolution and the estimation of remaining useful life (RUL) of components from models and data. The Prognostics algorithm uses data-driven models trained online based on data generated by the diagnosis module to make predictions about the fault progression. Such fault progressions are used to evaluate their impact on the mission objective (or system requirements). The reconfiguration module computes actuation configurations that respond to fault progressions.  We employ PDDL-based planning \cite{journals/corr/abs-1106-4561} enhanced with semantic attachments  \cite{semantic_attachements} to  efficiently compute optimal system configurations. The reconfigurator will be able to eliminate the impact of a system fault on mission objectives if the system is designed with insufficient actuation capacity. We use a tool-independent model representations, namely Functional Mockup Units (FMUs)  \cite{blochwitz2011functional} that are integrated into the three modules. FMUs are computational models of system models developed for simulation purposes.

We have designed our software implementation such that the three modules could
be deployed independently. We achieved such decoupling through the use of a
publish-subscribe messaging pattern, along with the design of a clearly-defined
API that enforces a consistent data representation throughout. The general
workflow being followed is depicted in Figure~\ref{fig:approach diagram}. The
diagnosis module subscribes to sensor measurements from the system, receives a
batched set of sensor data, and performs parameter estimation to compute a
diagnosis. In the event of ambiguity in the diagnosis, disambiguation actions
are published back into the messaging service and interpreted by actuators in
the system. The diagnosis module is able to horizontally scale across available
computational resources. Diagnoses are also published so that they are
accessible to downstream modules, such as the case of prognosis. After each
diagnosis, the prognosis module is triggered to update its online damage
progression model -- as well as update its RUL estimates -- using a non-linear
regression approach. Lastly, the reconfiguration module, which subscribes to
diagnosis and prognosis updates, is engaged to compute a configuration that
maximizes mission success based on the current system state and progression
estimate. While the reconfiguration module expects damage progression estimates,
it will use the fault state predicted by the diagnosis and assume a constant
progression in case the prognosis module is unavailable or unresponsive, for
added resilience.

\textit{Notations:} We use upper-case to denote random variables ($X$) and lower-case to denote a realization of a random variable ($x$). We use bold letters to denote vectors ($\boldsymbol{x}$). We mark the continuous time dependency by using the notation $X(t)$ and $x(t)$ for random processes and time-varying variables, respectively. To represent discrete time dependency, we use the notation $x(t_k) = x_k$, for time instants $t_k$.
A sequence of variables over time $\{x_k\}_{k=0}^K$ is denoted by $x_{0:K}$. We denote  the probability distribution function (p.d.f.) of a random variable $X$ by $f_{X}(x)$. We represent  the conditional p.d.f. of $X|Y$ by $f_{X|Y}(x|y)$. When there is no loss of clarity, to simplify the notation, we will omit the subscript notation of $f_{X|Y}(x|y)$, that is, we will use $f(x|y)$. We denote the expectation of the random variable $X$ by $\mathds{E}[X]$. Let $\mathcal{S}=\{s_i\}_{i=1}^n$ denote a set of elements. We denote by $s_{-i}$ the set $\mathcal{S}-\{s_i\}$.

\textit{Paper structure:} In Section \ref{problem_setup} we discuss the diagnosis, prognostics and reconfiguration problems and describe the system model we use to showcase our approaches. In Section \ref{diagnosis} we present the pervasive diagnosis approach and introduce an algorithm for disambiguating uncertain diagnosis solutions. In addition, we describe several methods for improving the scalability of the diagnosis algorithm to real time implementation. We describe our prognostics algorithm in Section \ref{prognostics}, where we show how the results of MBD are used to train regression models online that predict RUL. In Section \ref{sec:reconfiguration} we describe our planning-based approach to system reconfiguration, as an avenue to reduce the impact of faults on the system behavior. We showcase our approach to ensuring system resilience through end-to-end results in Section \ref{sec:end-to-end-results}, results that include the diagnosis, prognostics and reconfiguration modules.

\section{Problem setup}
\label{problem_setup}
In this section we present the Fuel system model and we formalize the diagnosis, prognostics and reconfiguration problems.  We conclude with our definition of resilience.
\subsection{System model}

The health monitoring and reconfiguration modules use physics-based
models to reason about the state of the system. The types of systems
we encounter include nonlinearities, discrete and algebraic constraints.
The mathematical model describing the behavior of the physical system is given by a (hybrid) differential algebraic equation (DAE) of the form
\begin{eqnarray}
\label{eq:10080953}
0 &=& F(\dot{\boldsymbol{X}}, \boldsymbol{X}, \boldsymbol{U};\boldsymbol{P}), \ \boldsymbol{X}(0) = \boldsymbol{X}_0\\
\label{eq:10080954}
\boldsymbol{Y} &=& h(\boldsymbol{X}, \boldsymbol{U};\boldsymbol{P})+\boldsymbol{V},
\end{eqnarray}
where $\boldsymbol{X}$ is the (stochastic) state of the system, $\boldsymbol{U}$ is the vector of inputs, $\boldsymbol{P}$ is the vector of model parameters, and $\boldsymbol{Y}$ is the vector of output measurements. The outputs are affected by the independent and identically distributed (i.i.d.), additive noise $\boldsymbol{V}$, assumed Gaussian with zero mean and covariance matrix $\Sigma_v$. The initial state $\boldsymbol{X}_0$ and the vector of parameters are vector-valued random variables with known prior distribution $f_{\boldsymbol{X}_0}$ and $f_{\boldsymbol{P}}$, respectively. For example, the p.d.f. of the vector of parameters can be Gaussian, with mean ${\boldsymbol{\bar{p}}}$ and covariance matrix $\Sigma_p$. The vector ${\boldsymbol{\bar{p}}}$ can be interpreted as the nominal value of the vector of system parameters, and matrix $\Sigma_p$ reflects the uncertainty in the nominal value.

\begin{figure}[htp!]
  \centering
  \includegraphics[width=0.96\linewidth]{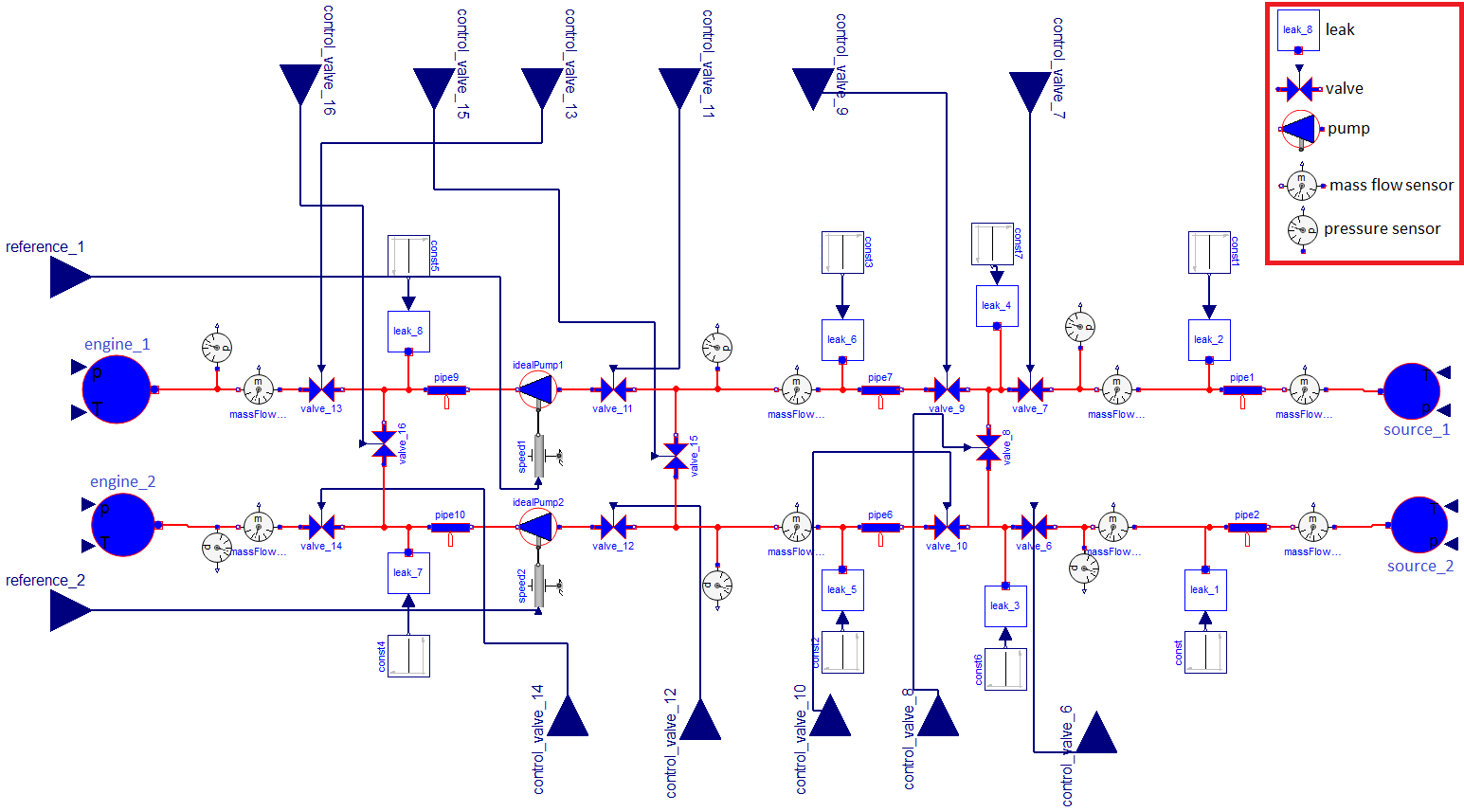}
  \caption{Modelica model of a fuel system.}
  \label{fig:fuel system}
\end{figure}

To illustrate the end-to-end health monitoring and reconfiguration framework we constructed a fuel system model. We use Modelica \cite{Fritzson15}
to model all our systems.  The model describes the fuel supply from two tanks to two engines through a series of pipes and valves. This model is a proxy for the fuel system model used in an autonomous ship.  We present the  Modelica block diagram of the fuel system in Figure \ref{fig:fuel system} The model uses components from the  {\tt Modelica.Thermal.FluidHeatFlow} library.
We model the tanks and the engines using ambient sources and sink components. They form two fuel lines that in nominal conditions operate independently. There are four valve components for each line that control the mass flow rate of fuel.  The two main lines that are connected through three cross-line valves, along the fuel lines. They ensure that the fuel can be re-routed from one line to another, as needed. Each valve is controlled by a continuous actuation signal, taking values in the interval [0, 1]: a 0 value means that the valve is completely closed, a 1 value means that the valve is completely open. There are a total of 11 control signals operating on the valves. Two fuel pumps, one for each line, move fuel from the tanks to the engines. The volumetric flow rate of each pump is determined by a control signal that sets its angular velocity.  We can measure fuel mass flow rates at 8 location along the fuel lines (see Figure \ref{fig:fuel system}). We can measure pressure at 4 locations. These pressure measurements, in addition to the mass flow rate measurements, will constitute boundary conditions for the components. 

The fuel system shown in Figure \ref{fig:fuel system} is affected by two types of faults: \textit{leaks} and \textit{stuck valves}. To include faulty behavior in the model, we augment the model with parameterized components that enable fault injection. This approach to fault modeling bulds on  \cite{Saha2014,Minhas2014}.
The leaks are modeled through specialized components that use (fault) parameters to set the severity of the faults. A 0 value means there is no leak, while a 1 value means the highest leak severity.  We have included 4 leak points on each line. The leak components are implemented using a valve connected to a sink component that sets ambient pressure. The fault parameter sets the opening of the valve. The stuck valve is modeled by setting the valve control signal and keeping it at a constant value determined by the associated fault parameter. Since the stuck faults are associated with valves, we have a total of 11 stuck faults. 

\subsection{Diagnosis}
\label{sec:diagnosis}
Broadly speaking there
are two approaches to the diagnosis problem: MBD and Machine Learning (ML). MBD methods require models and parameters
while ML approaches need a large amount of training data and extensive feature engineering. Since in our case we do have a system model but no failure data, we use MBD. In model-based approaches, the diagnosis
engine is provided with a model of the system, nominal values of the
parameters of the model and values of some of its inputs
and outputs. The main goal of a diagnosis engine is to determine from only this information the presence of a fault and to isolate it. There is a rich literature on model-based diagnosis results proposed independently by the artificial intelligence \cite{deKleer92} and control \cite{Gertler_1998},\cite{Isermann200571},\cite{Patton_2000} communities. Traditional model-based diagnosis approaches in the control communities include filters (e.g., Kalman filter \cite{Kalman}, particle filter \cite{Arulampalam02atutorial}), or optimization based-techniques that estimate parameters whose deviation from their nominal values indicate the presence of a fault. These methods rely on model simulations either for one sample period (Kalman and particle filters) or for some time horizon (optimization based).

We define $\mathcal{F}=\{1,2, \ldots, N\}$ as a set of ordered indices for the fault modes. Each fault mode has an associated scalar parameter $p_i$.  We denote the vector of all fault  parameters by $\boldsymbol{p}=[p_1, \ldots, p_{N}]$, and we assume it has a nominal value $\boldsymbol{\bar{p}} = [\bar{p}_1, \ldots, \bar{p}_{N}]$. In nominal conditions of the system, the parameter vectors remain close to $\boldsymbol{\bar{p}}$.  Under the single fault mode scenario, we model the fault mode as a random variable defined on the set of outcomes $\Omega=\{\omega_i\}_{i\in \mathcal{F}}$, where the event $\omega_i$ is defined as a discrepancy between a model parameter and its nominal value. In particular, $\omega_i=\{|p_i-\bar{p}_i|>\varepsilon_i, p_{-i} = \bar{p}_{-i}\}$, where $\varepsilon_i$ is a positive scalar. The scalar $\varepsilon_i$ depends on the measurement noise and the sensitivity of the behavior of the system to changes in parameter $p_i$. The fault magnitude is determined by estimating the value of the system parameter $p_i$. Given a sequence of input and output measurements over the time horizon $\tau$, the diagnosis problem consists of computing the conditional probability $\mathds{P}(\omega_i|\boldsymbol{y}_{0:\tau},\boldsymbol{u}_{0:\tau})$, together with the  estimation of the parameter $p_i$.

The optimization algorithm uses model simulations to update the fault parameters so that the simulated outputs match the measured outputs. We segment the output time series into non-overlapping windows, and for each window we update the fault parameters by solving least square problems. The choice of the size of the window depends on the time constants of the system dynamics. The size of the window must be correlated with the time needed by the optimization algorithm to generate a solution. If such a time is larger than the window size, we incur delays in generating diagnosis solutions. The least square optimization algorithm is expressed as a nonlinear program that optimizes for continuous fault parameters, i.e., the leak magnitudes. For discrete fault parameters, such algorithms are not applicable. We either use mixed-integer programming, approach that does not scale with the size of the discrete values taken by the fault parameters, or we relax the discrete requirement. As an alternative to the optimization-based approach, we could use filtering-based techniques, by considering the fault parameters as states. Since the fuel system model is nonlinear, the linear Kalman filter cannot be applied directly.  The extended Kalman filter \cite{1966ITAES...2..613M} requires artifacts not readily available such as the Jacobians of the state and measurements maps. Sampling-based filtering techniques, such as the particle filter \cite{Arulampalam02atutorial}, are computationally intensive since they require many sample points to propagate an accurate (possibly non-Gaussian) distribution of the state. The unscented Kalman filter \cite{1997SPIE.3068..182J} is a compromise between accuracy and computational effort and uses a set of sigma-points to approximate the distribution of the state. Qualitative diagnosis algorithms, such as analytical redundant relations (ARRs) \cite{STAROSWIECKI2000301,STAROSWIECKI2001687} can also be applied. They have the advantage that they do not require fault models, but they typically need more sensors to generate unambiguous diagnosis solution. Since in the case of the Fuel System the physics behind leaks is well understood,  we opted for a diagnosis solution based on fault models. Since the framework is intended to be deployed on an autonomous-ship, scalability of the framework to real-time execution is of utmost importance. We address this challenge by: (i) employing parallel execution of processes that compute diagnosis solutions, (ii) applying system decomposition to decouple the system into independent subsystems and diagnose them in parallel, (iii) using surrogate models to efficiently compute interventions to reduce diagnosis uncertainty, and (iv) improving the performance of the planner via model-specific heuristics, and reducing the complexity of the mixed discrete/continuous planning problem via temporal discretization with a finite time horizon.

%


\subsection{Prognostics}
Prognostics is the science of predicting the health condition of a system and/or its components using information about the past usage, current state, and future conditions. Past usage can be inferred from historical data of similar system operations. The current state is determined from sensor measurements, while future conditions are defined by the operational conditions under which the health of the system is evaluated. Prognostics predicts the future performance of a component by evaluating the deviation or degradation of a system from its expected normal operating conditions. The time until a component reaches a failure beyond which the system can no longer be used to meet desired performance is defined as the remaining useful life (RUL).

Similar to the diagnosis problem, data-driven and model-based approaches are used for prognostics. Data-driven prognostics  use pattern recognition and machine learning algorithms and models to detect changes in system states. Data driven models used for prognostics include autoregressive (AR) models, various types of neural networks (NNs) or neural fuzzy (NF) systems. Like diagnostics, there are two main data-driven strategies for prognostics: (i) modeling and learning the degradation of a model and extrapolating the RUL, and (ii) learning the RUL directly from data. Since running systems to failure is costly, the main challenge when using data-driven methods is the difficulty in obtaining enough. They require both a model of the system (dynamic equations that define the system behavior) and a model of the damage (or degradation) of a system/component as a function of environmental and operational conditions under which the system/component are operated. Examples of   physics-based degradation models include braking system wear based on the Archer's law \cite{doi:10.1063/1.1721448}), fatigue life model for ball bearings \cite{doi:10.1080/10402000108982420}, crack growth model \cite{10.1115/1.3656903}, or stochastic defect-propagation model \cite{LI2000747}. While physics-based degradation models exist for a number of applications, there is no formal approach to determine such models for all systems.

We use a hybrid prognostics approach: we combine a physics-based model of a system with data-driven model that learn online the degradation model and extrapolate the RUL. The data-driven model is trained based on the outputs generated by the diagnosis module. The diagnosis algorithm produces a sequence of parameter predictions $\{\hat{\boldsymbol{p}}_{k}\}_{k=0}^{\tau}$, where $t_{\tau}$ denotes the present time. We use this sequence to learn a dynamical model,
\begin{equation}
\label{eq:10122038}
\hat{\boldsymbol{p}}_k = G\left(\hat{\boldsymbol{p}}_{k-1}, \ldots, \hat{\boldsymbol{p}}_{k-n}, \boldsymbol{e}_k, \ldots, \boldsymbol{e}_{k-m}\right),
\end{equation}

\noindent for some positive integers $n, m$, where $\boldsymbol{e}_k$ stands for uncertainties and noise. This dynamical model represents the \textit{degradation model}. The RUL metric is defined in terms of component parameters or component variables. A component is declared failed when its parameter $p^i$ exceeds a range.  For some components only the maximum or minimum of the range matters.
In the case where a fault occurs if some maximum is passed, we define  the RUL for component $i$ as the random variable $K^i_{rul} = \min_k\{\hat{p}_{k,i}> p_{max,i}\}$, where $\hat{p}_{k,i}$ follows the stochastic dynamics predicted by the degradation model (\ref{eq:10122038}). The RUL metric can be defined in terms of system variables as well: $K^i_{rul} = \min_k\{x_{k,i}> x_{max,i}\}$, where $x_{k,i} = x_i(t_k)$ is a variable of component $i$. Predictions of $x_{k,i}$ are made using both the system model (\ref{eq:10080953})-(\ref{eq:10080954}) and the learned degradation model (\ref{eq:10122038}).\todo{Can we add when this fails.  i.e., what are the limitations.  Do we encounter them}

\subsection{Reconfiguration}
The Reconfiguration module is tasked with mitigating the fault's impact on operations and mission objectives. The mitigation is achieved by adjusting the relevant system's actuation controls. Reconfiguration uses as inputs information from the Diagnosis and Prognostics modules. These modules identify and localize faults, estimate their severity, and predict their evolution. Combined with an accurate dynamical model and mission specification, the reconfiguration module makes strategic adjustments based on the vessel's current and future condition, as well as its objectives and capabilities. Our approach for reconfiguration is \emph{Automated Planning} that transforms the reconfiguration task into a combinatorial search problem. In particular, we turn to hybrid (mixed discrete-continuous) planning, an expressive and feature-rich paradigm that allows for accurate modeling of realistic scenarios as planning domains. In such a setup, the system dynamics can include non-linear effects, exogenous activity, and concurrent events.

Planning models are normally defined using various levels of the Planning Domain Definition Language (PDDL)~\cite{mcdermott1998pddl}, a standardized modeling paradigm. PDDL is continuously evolving to enable reasoning with new concepts that bring AI Planning closer to solving real-world problems. The characteristics of the modeled system dictate the appropriate level of PDDL needed to accurately capture important features. PDDL+~\cite{fox2006modelling} was designed specifically to represent hybrid systems which exhibit both discrete and continuous behavior and introduced the concepts of continuous flows (processes) and discrete mode switches (events) to planning domains. Such domains are amongst the most advanced planning models and are notoriously challenging to solve due to exogenous activity, non-linear dynamics, and vast state space. However, PDDL+ and its expressive power are an important step in tackling complex real-world planning problems.

Various approaches to solving problems set in hybrid domains have been proposed in planning~\cite{cashmore2016compilation,bryce2015smt,coles2012colin}, as well as control and model-checking~\cite{cimatti1997planning,bogomolov2014planning}. However, all of the these techniques are limited in scale, model features, or dynamics.
To date, the only viable approach to planning in realistic hybrid domains is \emph{planning via discretization}~\cite{dellapenna2012universal,piotrowski2016heuristic,scala2016interval}. More formally, planning via discretization paradigm sees PDDL+ models translated into Finite State Temporal System (FSTS), an automaton which defines the state space and state transitions. An FSTS is a tuple $(S, s_0, \mathcal{A}, \mathcal{D}, F, T)$ where $S$ is a finite set of states (i.e., configurations the system can be in), $s_0 \in S$ the initial state, $\mathcal{A}$ is the set of actions that modify the states, $\mathcal{D} = \{0,\Delta t\}$ where $\Delta t$ is the discretized time step. $F : S \times \mathcal{A} \times \mathcal{D} \rightarrow S$ is the transition function generating successor states $s'$ (i.e., $F(s,a,d) = s'$), and $T$ is the finite temporal horizon.
A planning task focuses on finding a state $s_G$ that satisfies goal conditions, within the constraints and structure of the FSTS.

A solution to a PDDL+ planning problem in an FSTS is a plan, i.e., a trajectory or sequence of state, action, and duration tuples (i.e., $(s_t,a_t,d_t)$) that starts with the initial state $s_0$ and ends with the goal state $s_G$.

The vast majority of real-world scenarios are inherently hybrid in nature. We cast reconfiguration as a hybrid planning problem, encode the system in PDDL+, and exploit the \emph{planning-via-discretization} approach to solve the resulting combinatorial search task.

\subsection{Resilience}
\begin{figure}[htp!]
  \centering
  \includegraphics[width=0.8\linewidth]{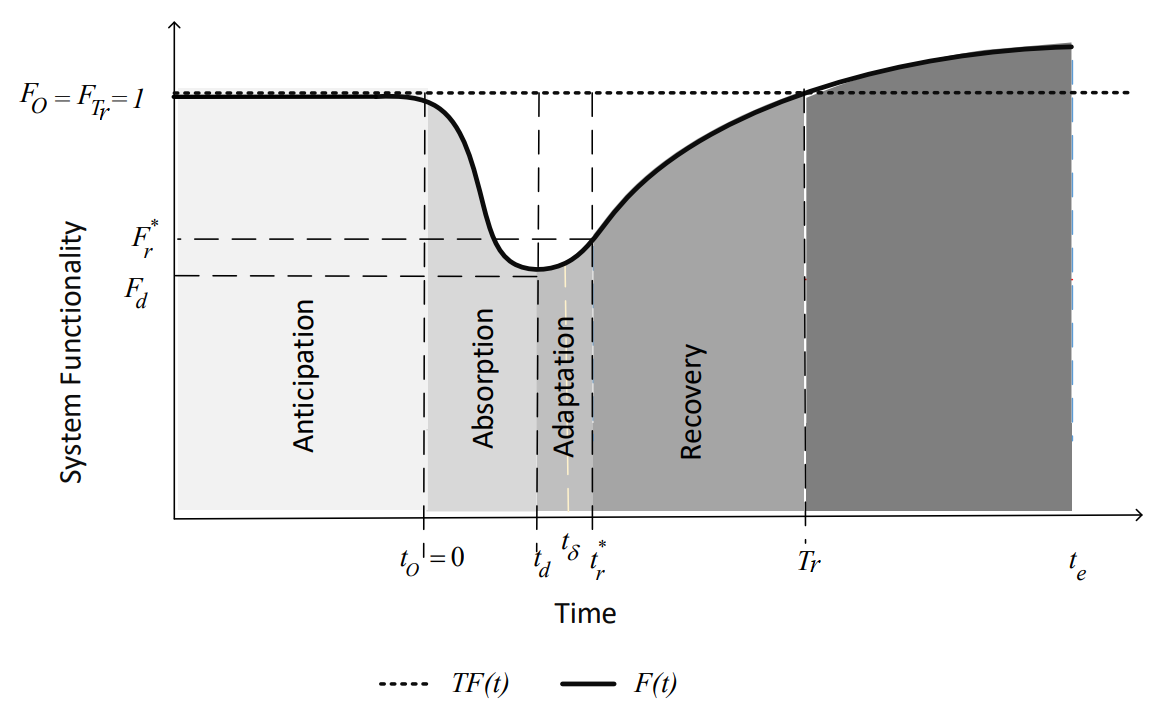}
  \caption{Phases that system undergoes in the face of an extreme event \cite{https://doi.org/10.1111/risa.13274,FRANCIS201490}.}
  \label{fig:system phases}
\end{figure}

It is difficult to define a  comprehensive definition of resilience. A qualitative definition of resilience inspired from power systems \cite{7842605,7105972} is:  \emph{a system is resilient if it is able to withstand, adapt and recover from unforseen events}. A more general, quantitative list of resilience metrics is proposed in \cite{https://doi.org/10.1111/risa.13274}, where resilience is defined with respect to abilities of resilience: anticipation, absorption, adaptation, and rapid recovery. These phases, defined with respect to extreme events,  are depicted in Figure \ref{fig:system phases}, where $TF(t)$ represents the expected normal functionality of the system, $F(t)$ is the true functionality of the system, $t_o$ is the time the extreme event happens, $t_d$ is the instant of the lowest functionality of the system, $t_r^*$ is the start of the recovery phase, while $T_r$ is the time the system re-enters the normal functionality.
The proposed resilience metrics defines with respect to the four abilities, are normalized with the simplest one being the fraction between the true functionality as the system goes through the four phases, and the nominal functionality, i.e., $\frac{\int_{0}^{T_r}F(t)dt}{\int_{0}^{T_r} TF(t)dt}$. A challenge with this type of metric is that it is defined with respect to the time $T$ at which the system recovers its functionality. However, a system may never recover its full functionality, yet still be able to perform its mission. In addition, the system may be in a continuous adaptation phase, as it continuously adapts and responds to faults; faults that evolve over time, as well. Remaining in the same spirit of resilience with respect to functionality, we define the system resilience with respect to a set of system requirements. In the case of the fuel system, the system requirements are (i) to ensure a prescribed mass flow rate at each of the two engines, and (2) to minimize the fuel losses, over some time horizon. The system should \emph{withstand} the effect of faults, by \emph{adapting} its operation (i.e., acting on the pumps and valves) to \emph{recover} from disruptions in supplying fuel to the engines. Let $\boldsymbol{y}^d = [{y}_1^{d},{y}_2^{d}]$ define the required mass flow rates at the engines, and let  $\boldsymbol{y}_k^{in}=[{y}_{1,k}^{in},{y}_{2,k}^{in}]$, $\boldsymbol{y}_k^{out}=[{y}_{1,k}^{out},{y}_{2,k}^{out}]$ define the measured fuel mass flow rates at the tanks and engines, respectively. Since the capacity of the fuel tanks is bounded, we bound the amount of fuel the tanks can supply, i.e.,  $\sum_{k=0}^T{y}_{i,k}^{in}\leq \beta_i$, where index $i$ designates tank $i$, and $\beta_i$ are positive scalars proportional to the tank volumes. We quantitatively define the fuel system resilience over a time horizon $T$ as the convex combination of two metrics that measure the deviation from the system requirements, i.e.,
\begin{equation}\label{eq:11050939}
  \mathcal{R}_T = 1-\alpha_1 \mathcal{J}^1_T - \alpha_2 \mathcal{J}^2_T,
\end{equation}
where $\alpha_i\geq 0$ and $\sum_i\alpha_i=1$, with
$\mathcal{J}^1_T = \frac{1}{(T+1)({y}_1^{d}+{y}_2^{d})}\sum_{i,k=0}^T|{y}_{i}^{d} - {y}_{i,k}^{out}|$, and  $\mathcal{J}^2_T = \frac{1}{\beta_1+\beta_2}\sum_{i,k=0}^T|{y}_{i,k}^{out} - {y}_{i,k}^{in}|$. The term $\mathcal{J}_T^1$ measures the deviations from the requires mass flow rates, while $J_T^2$ measures the fuel losses during the system operation over the time horizon $T$. Scalars $\alpha_1$ and $\alpha_2$ enables us to shift focus between the two requirements.  In nominal conditions, the terms $\mathcal{J}^1_T$ and $\mathcal{J}^2_T$ are zero, hence the system resilience is at its maximum, i.e., $\mathcal{R}_T = 1$. In the worst case, all fuel is lost and the system resilience is zero,  i.e., $\mathcal{R}_T = 0$. In Section \ref{sec:end-to-end-results} we will evaluate the resilience metric under different scenarios that demonstrates the effect of our proposed framework on the system resilience.

\section{Pervasive Diagnosis}
\label{diagnosis}
In this section we describe our optimization-based approach to MBD. 
For the cases when the transients are very fast and when the steady state does not depend on the initial conditions, the system model (\ref{eq:10080953})-(\ref{eq:10080954})  in its discrete form becomes a memoryless mathematical model:
\begin{equation}
\label{eq:10122057}
    \boldsymbol{Y}_k = h( \boldsymbol{U}_k;\boldsymbol{P})+\boldsymbol{V}_k,
\end{equation}
where we use upper cases to emphasize the stochastic nature of the system model. Given a sequence of output measurements $\{\boldsymbol{y}\}_{k=0}^{\tau}$ and inputs $\{\boldsymbol{u}\}_{k=0}^{\tau}$, we define the diagnosis problem as the computation of the p.d.f. $f(\boldsymbol{p}|\boldsymbol{y}_{0:\tau}, \boldsymbol{u}_{0:\tau})$, and the optimal estimator is given by $\hat{\boldsymbol{p}}_{\tau}=\mathds{E}[\boldsymbol{P}|\boldsymbol{Y}_{0:\tau}, \boldsymbol{U}_{0:\tau}]$. The maximum likelihood estimator is the solution of the optimization problem $\max_{\boldsymbol{p}} f(\boldsymbol{p}|\boldsymbol{y}_{0:\tau}, \boldsymbol{u}_{0:\tau})$, whose solution is $\hat{\boldsymbol{p}}_{\tau}$, when $f(\boldsymbol{p}|\boldsymbol{y}_{0:\tau}, \boldsymbol{u}_{0:\tau})$ is a Gaussian distribution.

Using a Bayesian approach, the $f(\boldsymbol{p}|\boldsymbol{y}_{0:\tau}, \boldsymbol{u}_{0:\tau})$ can be expressed as

$$f(\boldsymbol{p}|\boldsymbol{y}_{0:\tau}, \boldsymbol{u}_{0:\tau}) = \frac{\prod_{k=0}^{\tau}  f(\boldsymbol{y}_k|\boldsymbol{u}_{k},\boldsymbol{p})f(\boldsymbol{p})}{\int \prod_{k=0}^{\tau}  f(\boldsymbol{y}_k|\boldsymbol{u}_{k},\boldsymbol{p})f(\boldsymbol{p}) d\boldsymbol{p}},$$
where $f(\boldsymbol{p})$ is the prior distribution of the vector of parameters.

Under the input-output dynamics (\ref{eq:10122057}), the conditional p.d.f. $f(\boldsymbol{y}_k|\boldsymbol{u}_{k},\boldsymbol{p})$ is a Gaussian p.d.f. with mean ${\boldsymbol{\hat{y}}}_k = h( \boldsymbol{u}_k;\boldsymbol{p})$ and covariance matrix $\Sigma_v$. The mean vector ${\boldsymbol{\hat{y}}}_k$ is generated by simulating the system model, given the input $\boldsymbol{u}_k$ and the vector of parameters $\boldsymbol{p}$.
Under the additive Gaussian measurement noise assumption, we compute the solution of the maximum likelihood estimator by solving the following optimization problem:
\begin{equation}
    \label{eq:10132037}
    \min_{\boldsymbol{p}}\sum_{k=0}^{\tau} \left(\boldsymbol{y}_k-\boldsymbol{\hat{y}}_k\right)^T\Sigma_v^{-1}\left(\boldsymbol{y}_k-\boldsymbol{\hat{y}}_k\right) - \log f(\boldsymbol{p})
\end{equation}

There is no guarantee that problem (\ref{eq:10132037}) has a unique solution. The exact type of solution depends on the number of sensors and their placement. To address the solution non-uniqueness efficiently, we use the single fault scenario (even if we know there are multiple faults): we assume that at the diagnosis time only one  fault is responsible for the abnormal observation. This scenario is implemented by constraining the space where the vector of parameters $\boldsymbol{p}$ resides. In particular, we assume that $\boldsymbol{p} \in \{\boldsymbol{p}_1, \ldots \boldsymbol{p}_{N}\}$, where each vector $\boldsymbol{p}_i$ is of the form $\boldsymbol{p}_i=[\bar{p}_1, \ldots, \bar{p}_{i-1}, {p}_{i},\bar{p}_{i+1}, \ldots, \bar{p}_{N}]$. We define the fault mode random variable $\theta\in \mathcal{F}$, and we have that $\boldsymbol{p} = \boldsymbol{p}_i$ with probability $\mathds{P}(\theta =i)$. With this constraint on the  vector of parameters, the fault probability  in fault mode $i$ is given by
$$\mathds{P}(|p_i - \bar{p}_i|>\varepsilon_i) = 1 - \mathds{P}(|p_i - \bar{p}_i|\leq \varepsilon_i) = 1-\int_{\bar{p}_i - \varepsilon_i}^{\bar{p}_i + \varepsilon_i}f(p_i|\boldsymbol{y}_{0:\tau}, \boldsymbol{u}_{0:\tau})dp_i.$$
The optimal estimators $\hat{p}_i$ are computed by solving $N$  optimization problems of the form

\begin{equation}
    \label{eq:10141945}
    \min_{{p}_i}\sum_{k=0}^{\tau} \left(\boldsymbol{y}_k-\boldsymbol{\hat{y}}_k^i\right)^T\Sigma_v^{-1}\left(\boldsymbol{y}_k-\boldsymbol{\hat{y}}_k^i\right) - \log f({p}_i),
\end{equation}
where ${\boldsymbol{\hat{y}}}_k^i = h( \boldsymbol{u}_k;{p}_i)$.

The optimization problem (\ref{eq:10141945}) can be further simplified by assuming high uncertainty in the parameter vector $\boldsymbol{p}_i$ that can be expressed through a uniform distribution. Consequently, the term $\log f({p}_i)$ in the loss function of (\ref{eq:10141945}) can be neglected.

For non-linear systems, computing the  closed-form solution of $f({p}_i|\boldsymbol{y}_{0:\tau}, \boldsymbol{u}_{0:\tau})$ is intractable due to the denominator. We can approximate the p.d.f. $f({p}_i|\boldsymbol{y}_{0:\tau}, \boldsymbol{u}_{0:\tau})$ by taking advantage of the parameter samples that are generated as part of the optimization problem (\ref{eq:10141945}). At each iteration of the algorithm, the model is simulated at the current parameter value, and the model outputs are generated. We recall that each problem (\ref{eq:10141945}) is a scalar optimization problem. Let $\{{p}_i^m\}_{m=1}^M$ be a set of $M$ samples generated during the optimization process. Then we have the following approximation:
$$f({p}_i|\boldsymbol{y}_{0:\tau}, \boldsymbol{u}_{0:\tau}) \approx \frac{\sum_{m=1}^M\delta({p}_i-{p}_i^m)\prod_{k=0}^{\tau}  f(\boldsymbol{y}_k|\boldsymbol{u}_{k},{p}_i^m)}{\sum_{m=1}^M \prod_{k=0}^{\tau}  f(\boldsymbol{y}_k|\boldsymbol{u}_{k},{p}_i^m)}, $$
and the fault probability $\bar{q}_i=\mathds{P}(|p_i-\bar{p}_i|>\varepsilon_i)$ is given by:
\begin{equation}
  \label{eq:11010948}
  \bar{q}_i =1- \frac{\sum_{m\in \mathcal{M}_i}\prod_{k=0}^{\tau}  f(\boldsymbol{y}_k|\boldsymbol{u}_{k},{p}_i^m)}{\sum_{m=1}^M \prod_{k=0}^{\tau}  f(\boldsymbol{y}_k|\boldsymbol{u}_{k},{p}_i^m)},
\end{equation}
where $\mathcal{M}_i = \{m | p_i^m \in [\bar{p}_i - \varepsilon_i, \bar{p}_i + \varepsilon_i ]\}$.  We compute the normalized fault probability as $q_i = \frac{\bar{q}_i}{\sum_{j=1}^M\bar{q}_j}$.


\subsection{Optimization-based system diagnosis}

We use optimization algorithms to find an assignment to the vector of
fault parameters $\boldsymbol{p}$ such that our system model
simulation output is consistent with the observations. In
a \emph{single fault scenario} only one
fault can happen at a time. As such at most one off-nominal fault value $p_i$
is estimated in
$\boldsymbol{p} = [\bar{p}_1, \ldots, \bar{p}_{i-1},
{p}_{i},\bar{p}_{i+1}, \ldots, \bar{p}_{N}]$.  (In the multiple fault scenario, multiple $p_i$ are estimated.) We separately compute
the most consistent value for every $i$th fault parameter:
\begin{equation}
  \label{eq:diag_opt_single_fault}
  l_i = \min_{{p}_i}\sum_{k=0}^{\tau} \left(\boldsymbol{y}_k-\boldsymbol{\hat{y}}_k^i\right)^2
\end{equation}
\noindent where
$\boldsymbol{\hat{y}}_k^i$ is the system model simulation using the
off-nominal fault value $p_i$, or
$h(u_k; [\bar{p}_1, \ldots, \bar{p}_{i-1}, {p}_{i},\bar{p}_{i+1},
\ldots, \bar{p}_{N}])$. As Equation~(\ref{eq:diag_opt_single_fault})
depicts, we use the mean squared error between observation and
simulation as the loss function for optimization. Such a cost function results from (\ref{eq:10141945}) when assuming independent measurement noise, with identical variances. We only
report $p_i$ as a diagnosis if $|p_i - \bar{p}_i| >
\varepsilon_i$. The most likely set of diagnoses are ones in which
$l_i \leq l_{-i}$ holds. We can compute the confidence on a diagnosis using Equation (\ref{eq:11010948}).

Our system model simulation pipeline adheres to the Functional Mockup
Interface standard~\cite{blochwitz2011functional} for interfacing with
and executing physics-based models. Specifically, our diagnosis engine
takes as input objects called Functional Mockup Units (FMUs), which
are self-contained, \emph{natively} executable containers allowing for
the exchange and simulation of dynamic models. Multiple systems
modeling environments -- such as OpenModelica, Dymola, Simulink,
SimulationX, among others -- support exporting models FMU objects.
While forward simulations of FMUs are very efficient as these objects
can natively target a specific computing architecture, the do not natively posses constructs enabling automatic differentiation (AD).  As such, the
\emph{black-box}\footnote{The simulator does not have direct access to the model equations. It can only query the FMU.} nature of our simulation setup limits us to using
only \emph{gradient-free} optimization approaches to estimate the
fault vector. Our implementation leverages Scipy's\footnote{Scipy is a Python library used for scientific computing.} gradient-free
optimization algorithms, namely Powell. Alternatively, we can use numerical approximations to approximate gradients, but this approach does not scale with the number of variables.

Consider the single fault scenario.
In particular, it allows us to
parallelize the execution of Equation~(\ref{eq:diag_opt_single_fault})
across available computational resources, as each fault parameter
estimation procedure is independent from each other. We implemented a
multiprocessing architecture capable of running multiple
single-variable optimization procedures across a shared pool of
processes -- sized according to available resources, effectively
speeding up diagnosis. Each optimization procedure contains its own
FMU object handle which it uses for simulation and loss function
computation. An added advantage of this architecture is that it
allows us to recover from FMU simulation failures -- which can happen
due to multiple reasons, including numerical errors in the FMU's
solver, or invalid initial conditions -- without restarting the
optimization procedure, by simply spawning a new delegate simulation
process.  In the multiple fault case, we assign one process to each multiple fault we are interested in.

\begin{figure}[htp!]
  \centering
  \includegraphics[width=0.7\linewidth]{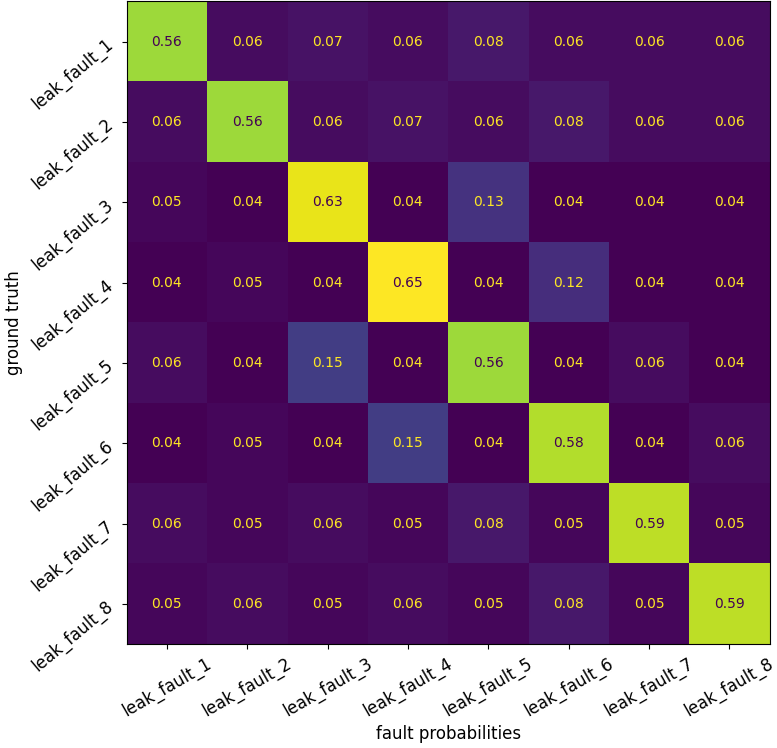}
  \caption{Fault probabilities versus the ground truth: low leak (fault parameter equals 0.25).}
  \label{fig:low_leak}
\end{figure}

\begin{figure}[htp!]
  \centering
  \includegraphics[width=0.7\linewidth]{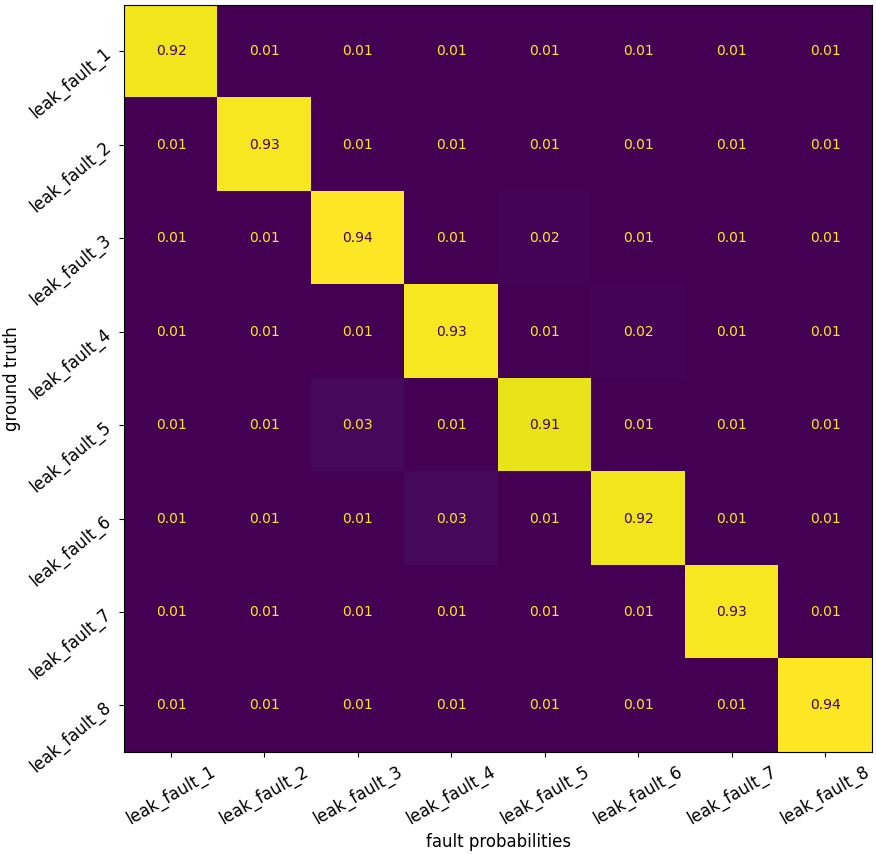}
  \caption{Fault probabilities versus the ground truth: medium leak (fault parameter equals 0.5).}
  \label{fig:medium_leak}
\end{figure}

\begin{figure}[htp!]
  \centering
  \includegraphics[width=0.7\linewidth]{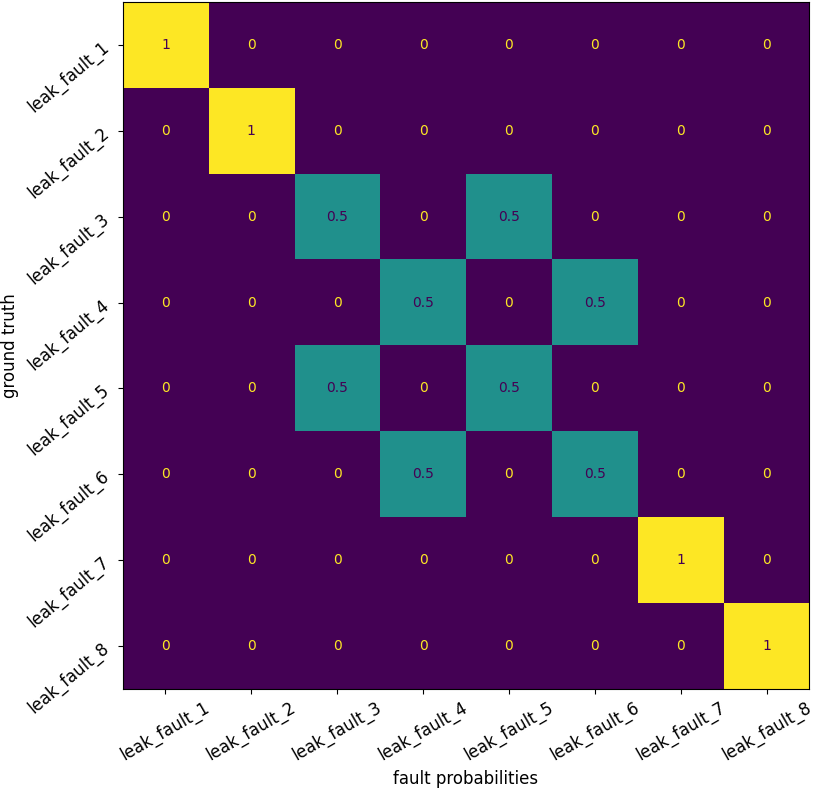}
  \caption{Fault probabilities versus the ground truth: high leak (fault parameter equals 0.75).}
  \label{fig:high_leak}
\end{figure}

We applied the parameter estimation based diagnosis algorithm for detecting leaks in the transportation of fuel from the tanks to the engines. We considered three leak magnitudes, \emph{low}, \emph{medium} and \emph{high}, which correspond to the fault parameter values 0.25, 0.5 and 0.75, respectively. These numbers relate to the amount of mass flow that is lost.  We tested the diagnosis algorithm for detecting leaks at six locations, and computed the fault probability of each leak location.  We used the Modelica model to generate data for each possible leak location and for the three fault magnitudes. We kept the same exogenous inputs (i.e., pump reference signals, valve openings) for each of the fault scenario. We recorded the mass flow rates at the eight measurement locations, to which measurement noise was added. We use these output measurements in the diagnosis algorithm. The diagnosis results are shown in Figures \ref{fig:low_leak}, \ref{fig:medium_leak} and \ref{fig:high_leak}. The labels corresponding to the rows represent the ground truth. The values of each row are the fault probabilities. The perfect result (i.e., the faults are identified with probability one) is the identity matrix. For low leaks, while there is some uncertainty in the fault likelihood, the probabilities corresponding to the ground truth are indeed dominant: hence, correct diagnosis are made. As the fault magnitude increase, the faults are easier to detect, as shown in the medium leak case. For the high leak case, we can no longer distinguish between {\tt leak\_fault\_3} and {\tt leak\_fault\_5}, and {\tt leak\_fault\_4} and {\tt leak\_fault\_6}. This result is not surprising: since due to the magnitude of the leak, all fuel is lost at the leak location. But since, there is no sensor in between the {\tt leak\_fault\_3} and {\tt leak\_fault\_5}, the diagnosis algorithm cannot pinpoint the exact location. Hence, the diagnosis is ambiguous. In the next section we show how we can solve the diagnosis disambiguation problem. Tables \ref{tab:small_leak}, \ref{tab:medium_leak} and \ref{tab:high_leak} show the parameter estimation results. The first column of the table identifies the active fault, while the first show that fault hypothesis. A correct result should show 0.25, 0.5 and 0.75 on the main diagonal of the three tables, respectively. In all cases the diagnosis algorithm estimates the fault magnitude correctly. The table values must be read in concert with the fault probabilities, i.e., we are interested in the parameter estimates for faults that have high probabilities. On a Dell Precision 3640 Tower (Intel Xeon CPU @ 4.1 GHz, 6 cores, 64 GB) the average time to generate a diagnosis is 1.5 sec.

\begin{table}[ht!]
\centering
\resizebox{\columnwidth}{!}{%
\begin{tabular}{|c|c|c|c|c|c|c|c|c|}
\hline
 & \textbf{leak\_fault\_1} & \textbf{leak\_fault\_2} & \textbf{leak\_fault\_3} & \textbf{leak\_fault\_4} & \textbf{leak\_fault\_5} & \textbf{leak\_fault\_6} & \textbf{leak\_fault\_7} & \textbf{leak\_fault\_8}\\
\hline
\textbf{leak\_fault\_1} & 0.25 & 0.08 & 0.19 & 0.04 & 0.2 & 0.05 & 0.17 & 0.05\\
\hline
\textbf{leak\_fault\_2} & 0.04 & 0.25 & 0.08 & 0.19 & 0.08 & 0.2 & 0.02 & 0.17\\
\hline
\textbf{leak\_fault\_3} & 0.22 & 0.09 & 0.25 & 0.03 & 0.25 & 0.02 & 0.17 & 0.05\\
\hline
\textbf{leak\_fault\_4} & 0.0 & 0.22 & 0.0 & 0.25 & 0.0 & 0.25 & 0.0 & 0.17\\
\hline
\textbf{leak\_fault\_5} & 0.22 & 0.0 & 0.24 & 0.0 & 0.25 & 0.0 & 0.21 & 0.01\\
\hline
\textbf{leak\_fault\_6} & 0.0 & 0.22 & 0.0 & 0.24 & 0.0 & 0.25 & 0.0 & 0.21\\
\hline
\textbf{leak\_fault\_7} & 0.19 & 0.0 & 0.16 & 0.04 & 0.21 & 0.04 & 0.25 & 0.07\\
\hline
\textbf{leak\_fault\_8} & 0.0 & 0.19 & 0.01 & 0.15 & 0.0 & 0.21 & 0.0 & 0.25\\
\hline
\end{tabular}
}
\caption{Leak fault parameter estimation for the \emph{small} leak case. For a correct estimation, the diagonal values should correspond to 0.25.}
\label{tab:small_leak}
\end{table}

\begin{table}[ht!]
\centering
\resizebox{\columnwidth}{!}{%
\begin{tabular}{|c|c|c|c|c|c|c|c|c|}
\hline
 & \textbf{leak\_fault\_1} & \textbf{leak\_fault\_2} & \textbf{leak\_fault\_3} & \textbf{leak\_fault\_4} & \textbf{leak\_fault\_5} & \textbf{leak\_fault\_6} & \textbf{leak\_fault\_7} & \textbf{leak\_fault\_8}\\
\hline
\textbf{leak\_fault\_1} & 0.5 & 0.0 & 0.43 & 0.0 & 0.45 & 0.0 & 0.4 & 0.0\\
\hline
\textbf{leak\_fault\_2} & 0.19 & 0.5 & 0.12 & 0.44 & 0.08 & 0.45 & 0.0 & 0.41\\
\hline
\textbf{leak\_fault\_3} & 0.47 & 0.0 & 0.5 & 0.0 & 0.5 & 0.0 & 0.41 & 0.0\\
\hline
\textbf{leak\_fault\_4} & 0.0 & 0.47 & 0.0 & 0.5 & 0.0 & 0.5 & 0.17 & 0.41\\
\hline
\textbf{leak\_fault\_5} & 0.47 & 0.18 & 0.48 & 0.0 & 0.5 & 0.03 & 0.46 & 0.2\\
\hline
\textbf{leak\_fault\_6} & 0.0 & 0.47 & 0.12 & 0.48 & 0.0 & 0.5 & 0.0 & 0.46\\
\hline
\textbf{leak\_fault\_7} & 0.43 & 0.13 & 0.4 & 0.15 & 0.46 & 0.13 & 0.5 & 0.01\\
\hline
\textbf{leak\_fault\_8} & 0.0 & 0.43 & 0.0 & 0.39 & 0.0 & 0.46 & 0.0 & 0.5\\
\hline
\end{tabular}
}
\caption{Leak fault parameter estimation for the \emph{medium} leak case. For a correct estimation, the diagonal values should correspond to 0.5.}
\label{tab:medium_leak}
\end{table}

\begin{table}[ht!]
\centering
\resizebox{\columnwidth}{!}{%
\begin{tabular}{|c|c|c|c|c|c|c|c|c|}
\hline
 & \textbf{leak\_fault\_1} & \textbf{leak\_fault\_2} & \textbf{leak\_fault\_3} & \textbf{leak\_fault\_4} & \textbf{leak\_fault\_5} & \textbf{leak\_fault\_6} & \textbf{leak\_fault\_7} & \textbf{leak\_fault\_8}\\
\hline
\textbf{leak\_fault\_1} & 0.75 & 0.11 & 0.7 & 0.1 & 0.7 & 0.04 & 0.67 & 0.0\\
\hline
\textbf{leak\_fault\_2} & 0.01 & 0.75 & 0.01 & 0.7 & 0.01 & 0.7 & 0.15 & 0.67\\
\hline
\textbf{leak\_fault\_3} & 0.75 & 0.04 & 0.75 & 0.1 & 0.75 & 0.13 & 0.72 & 0.12\\
\hline
\textbf{leak\_fault\_4} & 0.2 & 0.75 & 0.17 & 0.75 & 0.19 & 0.75 & 0.19 & 0.72\\
\hline
\textbf{leak\_fault\_5} & 0.75 & 0.2 & 0.75 & 0.14 & 0.75 & 0.12 & 0.72 & 0.0\\
\hline
\textbf{leak\_fault\_6} & 0.14 & 0.75 & 0.17 & 0.75 & 0.2 & 0.75 & 0.22 & 0.72\\
\hline
\textbf{leak\_fault\_7} & 0.75 & 0.05 & 0.75 & 0.08 & 0.75 & 0.08 & 0.75 & 0.03\\
\hline
\textbf{leak\_fault\_8} & 0.1 & 0.75 & 0.06 & 0.75 & 0.1 & 0.75 & 0.21 & 0.75\\
\hline
\end{tabular}
}
\caption{Leak fault parameter estimation for the \emph{high} leak case. For a correct estimation, the diagonal values should correspond to 0.75.}
\label{tab:high_leak}
\end{table}

\subsection{Diagnosis disambiguation}
\label{sec:disambiguation}
\begin{figure}[htp!]
  \centering
  \includegraphics[width=0.45\linewidth]{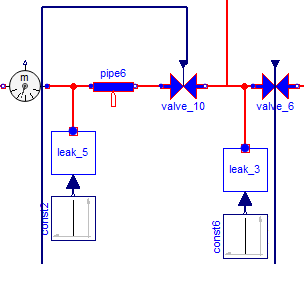}
  \caption{Local view of the fuel system, focusing on leaks that cannot be differentiated.}
  \label{fig:fuel system zoomed in}
\end{figure}

In an ideal scenario, the sensor measurements include sufficient information to distinguish among the fault modes. However, as seen in Figure \ref{fig:high_leak}, this is not always the case.  Figure \ref{fig:fuel system zoomed in} depicts a localized view of the fuel system, where we focus on {\tt leak\_fault\_3} and {\tt leak\_fault\_5}.
Given the two mass flow rate sensors ({\tt massFlowSensor\_4} and {\tt massFlowSensor\_6}) and the nominal inputs to the valves and the pumps, the diagnosis engine is unable to provide a precise diagnosis when the leaks are significant. We have an \textit{ambiguous} diagnosis solution. We define the ambiguity as two or more faults having similar probabilities. This result is not surprising since there is no sensor between the two leaks. We can address this weakness in two ways: (i) add more sensors, or (ii) elicit richer information from sensors through targeted system excitation. Since the first option entails changing the physical system, we use the second approach: we design valve and pump inputs so that when applied to the physical system, they generate sufficient information in the measurements. The new information is used for diagnosis disambiguation. The diagnosis result generates fault probabilities $q_i$ and  fault parameter estimations $\hat{p}_i$.  Assume $\mathcal{A}\subset \mathcal{F}$ is the set of ambiguous diagnoses, with corresponding fault parameter estimates $\{\hat{p}_i\}_{i\in \mathcal{A}}$. Our objective is to design a sequence of inputs $\{\boldsymbol{u}_k\}_{k=0}^{\tau}$ that maximizes the Euclidean distance between the system outputs that correspond to the ambiguous fault modes. We achieve this objective by solving the following optimization problem:
\begin{eqnarray}
\label{eq:10141457}
\min_{\boldsymbol{u}_0, \ldots, \boldsymbol{u}_{\tau}} & & -\sum_{i>j} \sum_{k=0}^{\tau}\|\boldsymbol{\hat{y}}_k^i - \boldsymbol{\hat{y}}_k^j\|^2\\
\label{eq:10141458}
\textmd{subject to:} &  &\boldsymbol{\hat{y}}_k^i = h(\boldsymbol{u}_k; \hat{p}_i), i\in \mathcal{A},\\
\label{eq:10141459}
 &  & \boldsymbol{u}_k\in \mathcal{U}, \forall k \in \{0, \ldots, \tau\},
\end{eqnarray}
where $\mathcal{U}$ is a set constraining the inputs that can be applied to the system. For example, in the case of the fuel system model, the valve inputs are constrained to take values in the interval $[0, 1]$.
At a first glance, it may appear that the choice for the disambiguation cost function is arbitrary; it is not. We recall that the key term for computing $f(p_i|\boldsymbol{y}_{0:\tau}, \boldsymbol{u}_{0:\tau})$ is the product of conditional p.d.f.s $\prod_{k=0}^{\tau}  f(\boldsymbol{y}_k|\boldsymbol{u}_{k},\boldsymbol{p})$. To evaluate each conditional p.d.f. $f(\boldsymbol{y}_k|\boldsymbol{u}_{k},p_i)$, we need to evaluate the quadratic term $\left(\boldsymbol{y}_k - \boldsymbol{\hat{y}}_k\right)^T\Sigma_v^{-1} \left(\boldsymbol{y}_k - \boldsymbol{\hat{y}}_k\right)$.  Let $i$ be the true fault mode, with $\boldsymbol{y}^i_k = \boldsymbol{\hat{y}}_k^i + \boldsymbol{v}_k$ the output measurements expressed in term of the simulated output $\boldsymbol{\hat{y}}_k^i$ and a realization of the measurement noise $\boldsymbol{v}_k$. For each ambiguous fault mode $j$, the quadratic expression $\left(\boldsymbol{y}_k - \boldsymbol{\hat{y}}_k\right)^T\Sigma_v^{-1} \left(\boldsymbol{y}_k - \boldsymbol{\hat{y}}_k\right)$ becomes $\left(\boldsymbol{\hat{y}}_k^i + \boldsymbol{v}_k - \boldsymbol{\hat{y}}_k^j\right)^T\Sigma_v^{-1} \left(\boldsymbol{\hat{y}}_k^i + \boldsymbol{v}_k - \boldsymbol{\hat{y}}_k^j\right)$. For the ground truth case $i=j$, the previous expression becomes $ \boldsymbol{v}_k^T\Sigma_v^{-1} \boldsymbol{v}_k$, which is the smallest quantity that we can get under the additive noise assumption. The smaller the quantity $\left(\boldsymbol{y}_k - \boldsymbol{\hat{y}}_k\right)^T\Sigma_v^{-1} \left(\boldsymbol{y}_k - \boldsymbol{\hat{y}}_k\right)$ is for all $k$, the closer to 1 the conditional p.d.f. $f(p_i|\boldsymbol{y}_{0:\tau}, \boldsymbol{u}_{0:\tau})$ becomes. Inversely, the larger $\left(\boldsymbol{y}_k - \boldsymbol{\hat{y}}_k\right)^T\Sigma_v^{-1} \left(\boldsymbol{y}_k - \boldsymbol{\hat{y}}_k\right)$ for all $k$ becomes, the closer to zero $f(p_j|\boldsymbol{y}_{0:\tau}, \boldsymbol{u}_{0:\tau})$ gets. Hence by solving (\ref{eq:10141457}), we in fact maximize the conditional p.d.f. $f(p_i|\boldsymbol{y}_{0:\tau}, \boldsymbol{u}_{0:\tau})$ for the ground truth fault mode $i$, while in all the other fault modes $j$, $f(p_j|\boldsymbol{y}_{0:\tau}, \boldsymbol{u}_{0:\tau})$ becomes smaller.

The disambiguation problem minimizes the diagnosis uncertainty. The information theory entropy, defined as  $-\sum_{i\in \mathcal{A}}q_i \log q_i$, where $ q_i$ are the probabilities of the ambiguous fault modes, is a metric that measures the diagnosis uncertainty.  Indeed, solving (\ref{eq:10141457}) brings the probabilities $q_i$ closer to zero or one, hence minimizing the entropy.

\subsection{Improving the scalability of the diagnosis algorithm}
Solving the diagnosis problem requires repeated simulations of the system model. Such a problem can be viewed as a black-box optimization since we can not analytically compute the gradients of the loss function; gradients that depend on the system variables. The black-box optimization limits the access to gradient-free optimization algorithms (e.g., Bayesian optimization, Powell, Simplex). Such algorithms scale poorly with the number of optimization variables. We discuss several avenues for improving the diagnosis scalability to real-time implementation.


\subsubsection{System decomposition}
\begin{figure}
\begin{subfigure}{.5\textwidth}
  \centering
  \includegraphics[width=\textwidth]{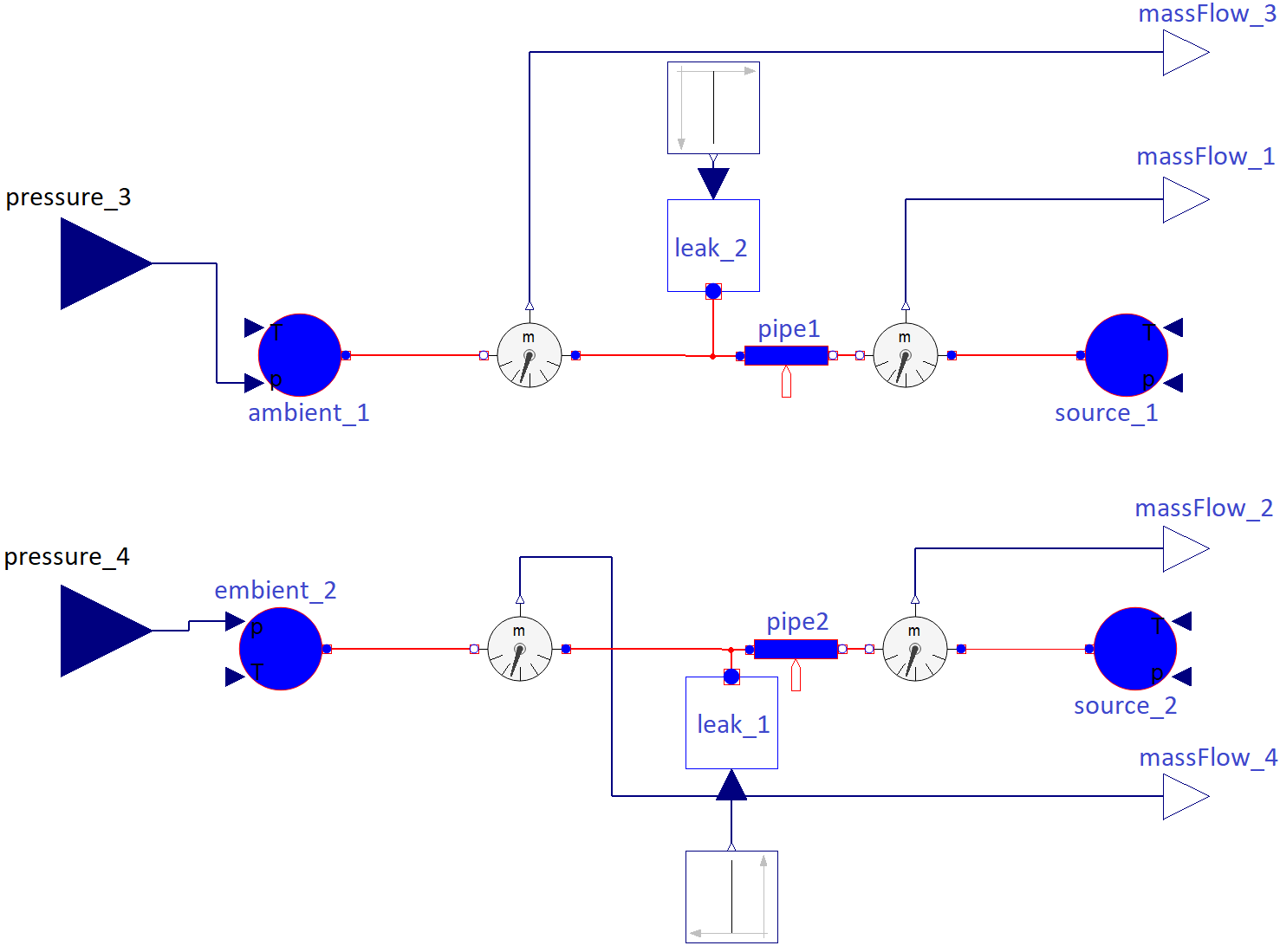}
  \caption{Subsystem 1.}
  \label{fig:subsystem 1}
\end{subfigure}%
\begin{subfigure}{.5\textwidth}
  \centering
  \includegraphics[width=\textwidth]{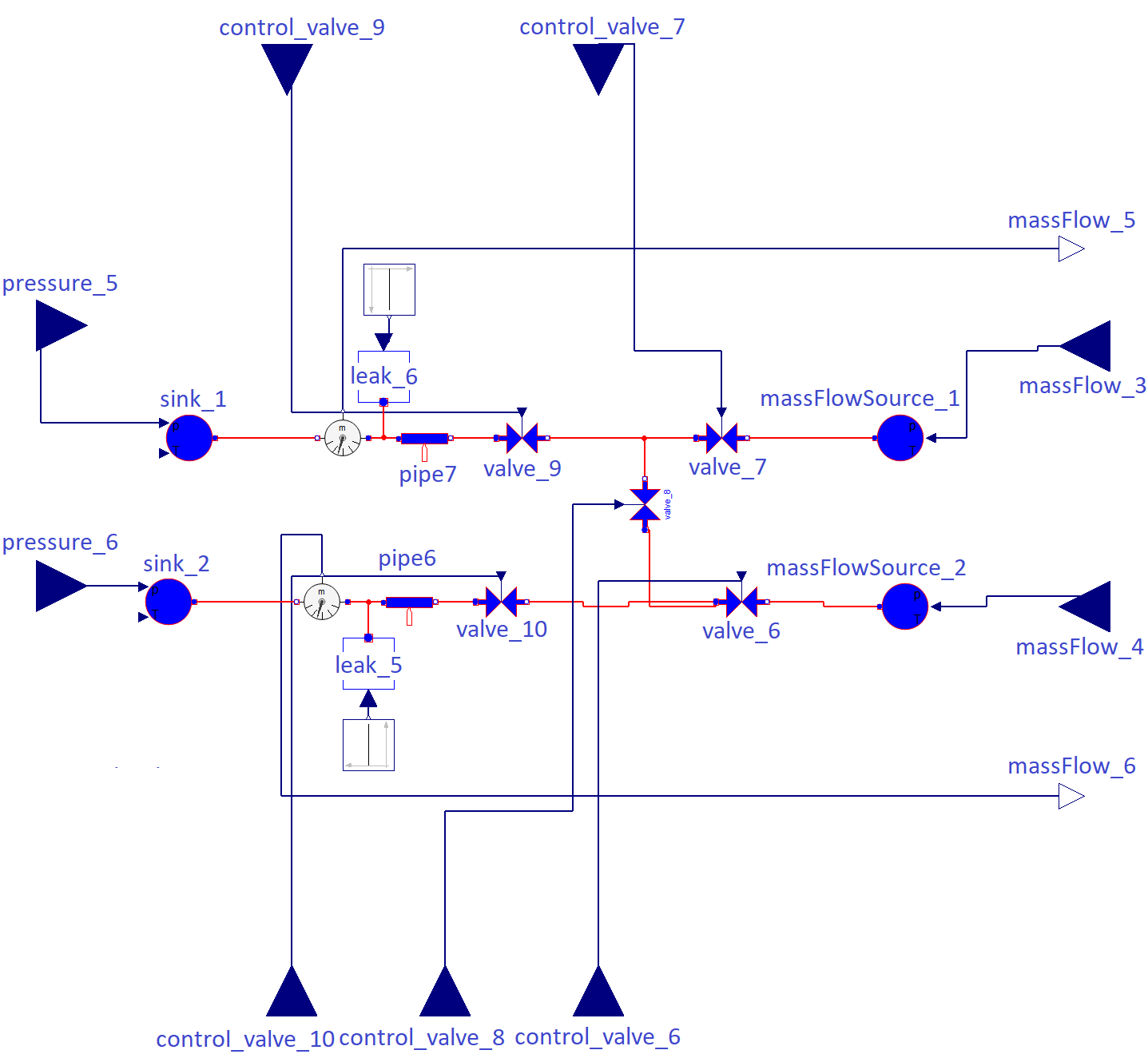}
  \caption{Subsystem 2.}
  \label{fig:subsystem 2}
\end{subfigure}\\
\begin{subfigure}{.5\textwidth}
  \centering
  \includegraphics[width=\textwidth]{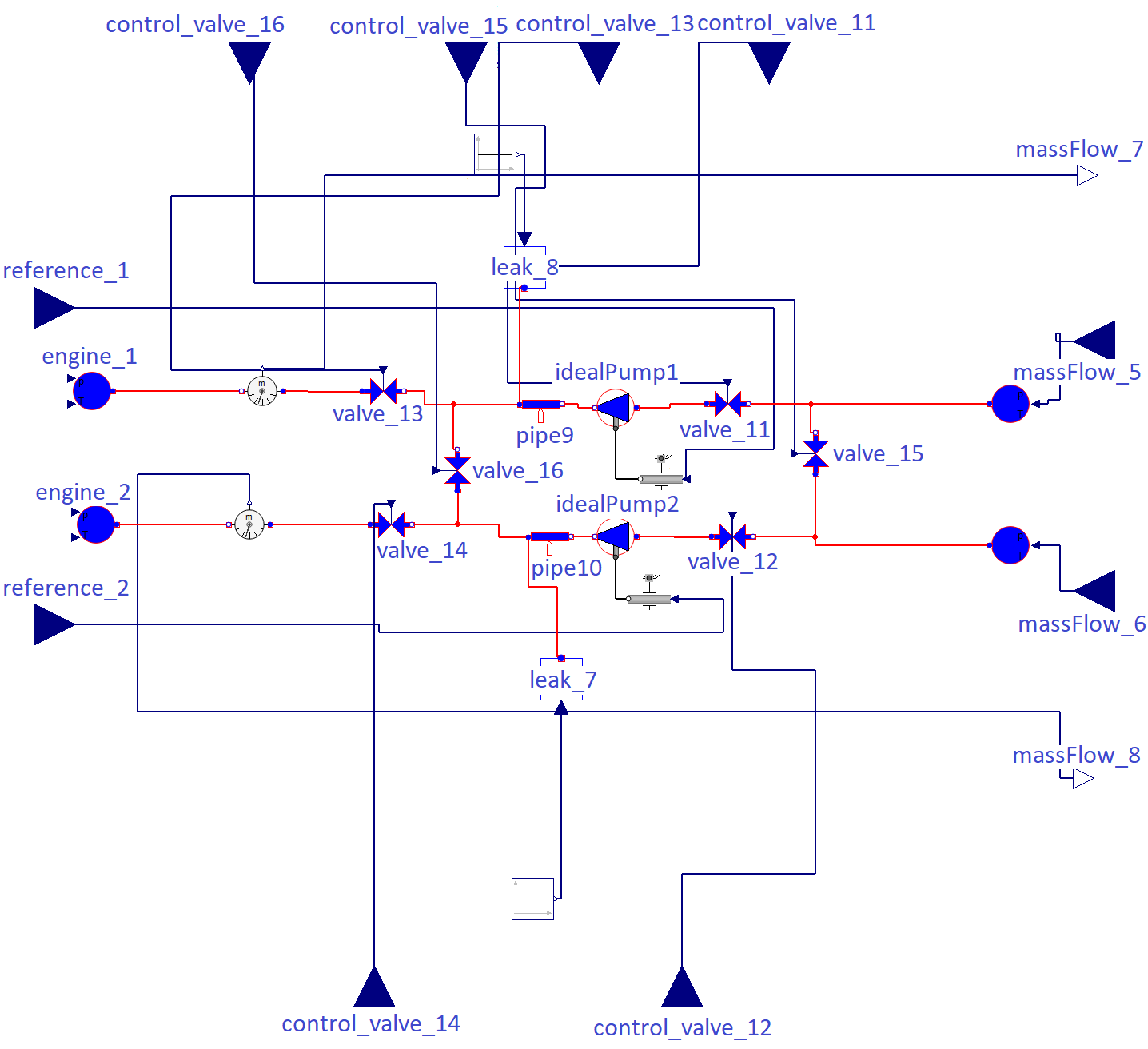}
  \caption{Subsystem 3}
  \label{fig:subsystem 3}
\end{subfigure}
\caption{Decomposition of the fuel system into three subsystems. Each subsystem can be diagnosed independently.}
\label{fig:fuel system decomposition}
\end{figure}

\begin{figure}[htp!]
  \centering
  \includegraphics[width=0.8\linewidth]{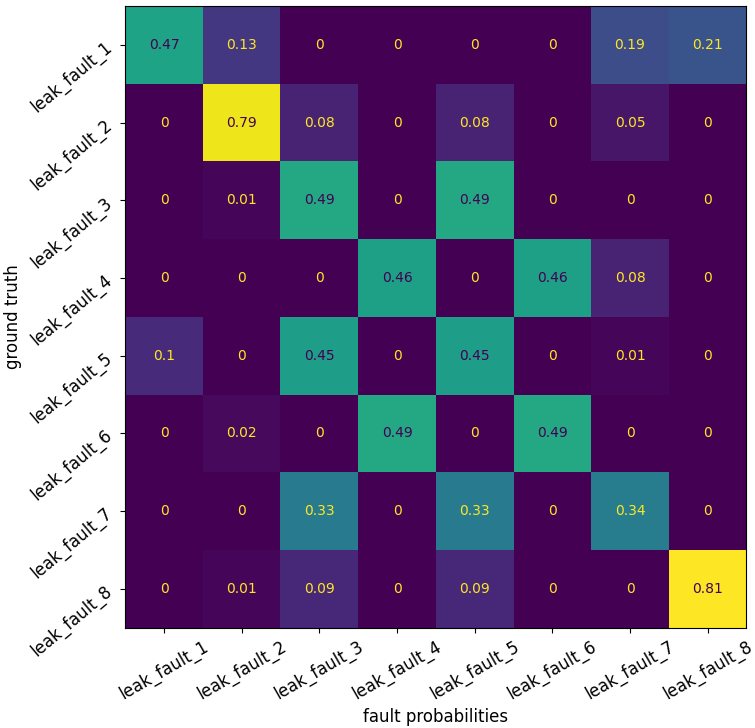}
  \caption{Fault probabilities versus the ground truth: low leak (fault parameter equals 0.25).}
  \label{fig:subsystem low_leak}
\end{figure}

\begin{figure}[htp!]
  \centering
  \includegraphics[width=0.8\linewidth]{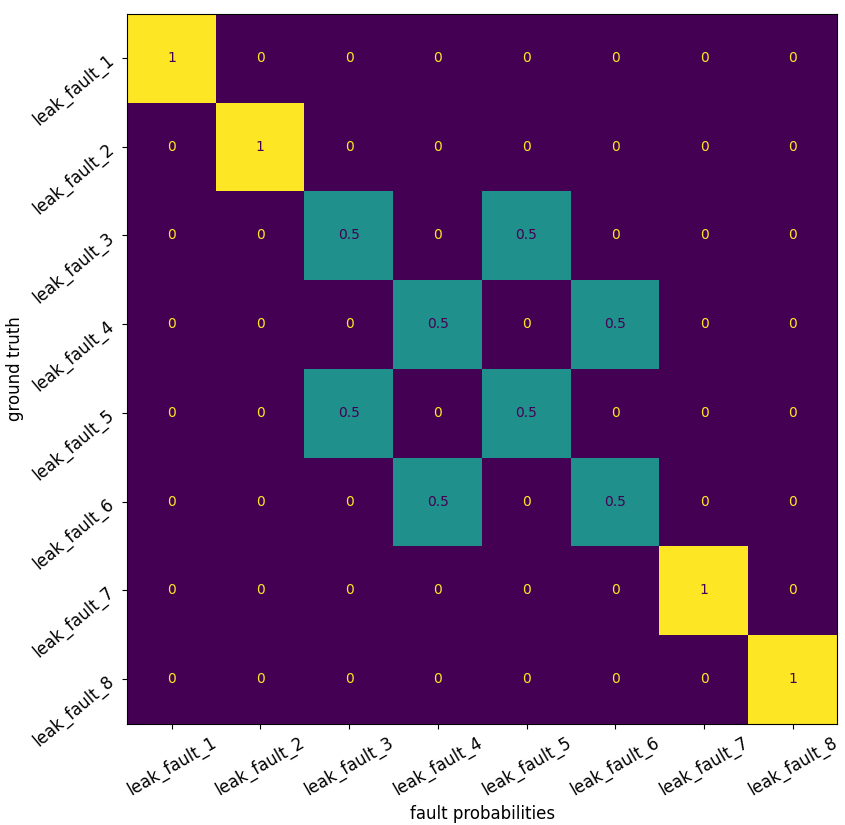}
  \caption{Fault probabilities versus the ground truth: medium leak (fault parameter equals 0.5).}
  \label{fig:subsystem medium_leak}
\end{figure}

\begin{figure}[htp!]
  \centering
  \includegraphics[width=0.8\linewidth]{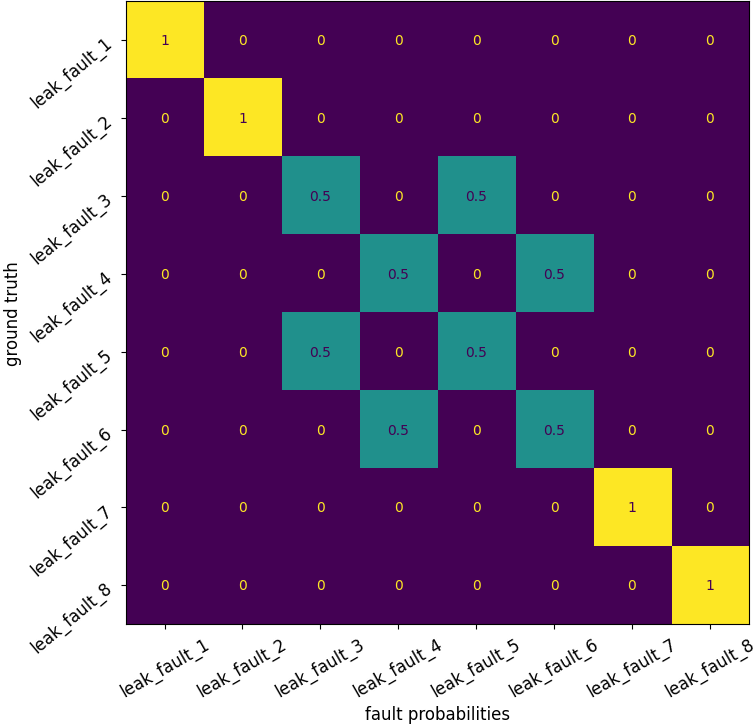}
  \caption{Fault probabilities versus the ground truth: high leak (fault parameter equals 0.75).}
  \label{fig:subsystem high_leak}
\end{figure}

Simpler models are typically easier to simulate. Hence, one approach to generate a diagnosis solution faster us to split the system into subsystems that are diagnosed independently. The system decomposition depends on the number of sensors and their locations. When simulating the system model, we solve a system of nonlinear equations to determine the system variables. The actual solving  is preceded by a set of operations that brings the system of equations to its simplest form: the block lower triangular (BLT) form \cite{XML_DAE}. In this form, the causality relations between system variables are emphasized: what variables are needed to compute other variables. When separating a subsystem from the rest of the system, we remove input variables that are needed to compute the values of the subsystem variables. We provide the removed input variables through sensor measurements that effectively become inputs for the subsystem. The subsystem can be viewed as an independent system  with boundary conditions set through sensor measurements. The proposed decomposition is not unique. We could have used pressure measurements only as boundary conditions. The only requirement is to replace through sensor measurements the variables that can no longer be computed after splitting off the subsystem from the rest of the system. Once the system is decomposed, we can diagnose the resulting subsystems independently. Since each subsystem is less complex, diagnosis time decreases.  Figure \ref{fig:fuel system decomposition} shows the decomposition of the fuel system into three subsystems. Subsystem 1 uses as inputs (boundary conditions) pressure measurements {\tt pressure\_3} and {\tt pressure\_4} that feed into ambient sources. With these inputs we can compute the mass flow rates as measured by {\tt massFlowSensor\_3} and {\tt massFlowSensor\_4}. Subsystem 2 uses as inputs both mass flow rates measured from Subsystem 1 ({\tt massFlowSensor\_3} and {\tt massFlowSensor\_4}) and pressures {\tt pressure\_5} and {\tt pressure\_6}, and computes as outputs mass flow rates {\tt massFlowSensor\_5} and {\tt massFlowSensor\_6}. Subsystem 3 uses as inputs the measurements provided by {\tt massFlowSensor\_5} and {\tt massFlowSensor\_6} and computes the mass flow rates at {\tt massFlowSensor\_7} and {\tt massFlowSensor\_8}.
We modified the diagnosis algorithm to accommodate independent diagnosis of the three subsystem. The algorithms run parallel processes that estimate the fault parameters. Unlike the case where the entire system is considered, each subsystem diagnosis considers a smaller number faults: 2 for subsystem 1, 4 for subsystem 2, and 2 for subsystem 4. There are two advantages of this approach: (i) shorter diagnosis time (as the model size decreases, typically the simulation time decreases as well), and (ii) distributivity of computational resources (subsystem diagnosis can be run on separate hardware devices). The diagnosis results are shown in Figures \ref{fig:subsystem low_leak}, \ref{fig:subsystem medium_leak} and \ref{fig:subsystem high_leak}. Unlike the case where the entire system is diagnosed, the algorithm is not able to disambiguate between {\tt leak\_fault\_3} and {\tt leak\_fault\_5}, and {\tt leak\_fault\_4} and {\tt leak\_fault\_6} even for the small leak faults. The reason for this result is the fact that the diagnosis algorithms applied to each subsystem make decisions based on subsystem measurement only. Hence, they do not take into account how setting parameters in one subsystem affects the output predictions of the other subsystems. In case the diagnosis results are ambiguous, we design custom inputs for disambiguation as described in Section \ref{sec:disambiguation}, and re-run the diagnosis algorithms with the new sensor measurements.

\subsubsection{Surrogate models}

 When solving optimization problems, gradient-free algorithms scale poorly with the number of optimization variables. The reason we are forced to use gradient-free methods is that the FMU representation of the physics-based model is a black box from the optimization perspective. Numerical approximation of gradients does not solve the problem: we would have to simulate the FMU for at least the number of the optimization variables. To use gradient-based optimization methods we have to \emph{open} the model. We alleviate the gradient computation problem by constructing approximation models, known as \emph{surrogate models}. Such models are emulators that mimic the behavior of the original model while enabling the evaluation of gradients at a cheaper computational cost. Surrogate models are constructed using a data-driven approach they focus on the input-output behavior, disregarding the inner working of the simulation code. We choose a particular type of surrogate models: a type that includes construct enabling automatic differentiation. Such models are typical in deep learning platforms such a Pytorch \cite{paszke2017automatic}. Our interest in gradient-based algorithms stems from the fact that, unlike gradient free algorithm, they scale linearly with the number of optimization variables.

\begin{figure}[htp!]
  \centering
  \includegraphics[width=0.5\linewidth]{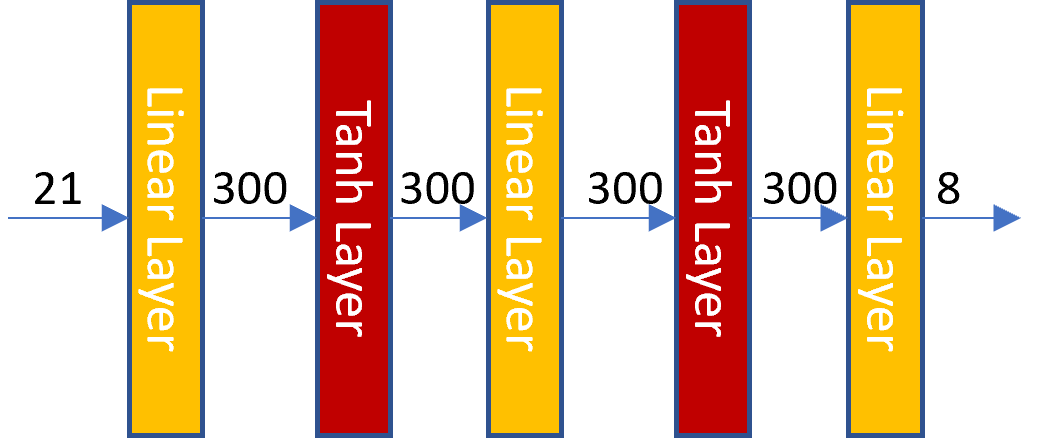}
  \caption{Surrogate model: neural network architecture}
  \label{fig:surrogate model}
\end{figure}

To improve the \emph{time} efficiency of the algorithm to resolve the diagnosis ambiguities, we designed and trained a neural network (NN) based surrogate model for the physics-based model. The structure of the model is shown in Figure \ref{fig:surrogate model}.
The model was implemented in Pytorch and has two hidden layers of size 300, with {\tt tanh} as activation function. We use dense NNs since we model a memoryless, input-output model. The surrogate model has as inputs the exogenous inputs and parameters of the physics-based model, and as outputs the sensor measurements of the physics-based model. Hence, the surrogate model preserves the functionality of the physics-based model. The training data is generated by simulating the physics-based model. For a large number of inputs and outputs, we have a combinatorial explosion in the size of the training data. However, the training data generation is done off-line, hence time constraints are less relevant. When independently diagnosing subsystems, we train surrogate modes for smaller models, with fewer inputs and parameters. Hence, the training data generation becomes more manageable.

\begin{figure}[htp!]
  \centering
  \includegraphics[width=0.8\linewidth]{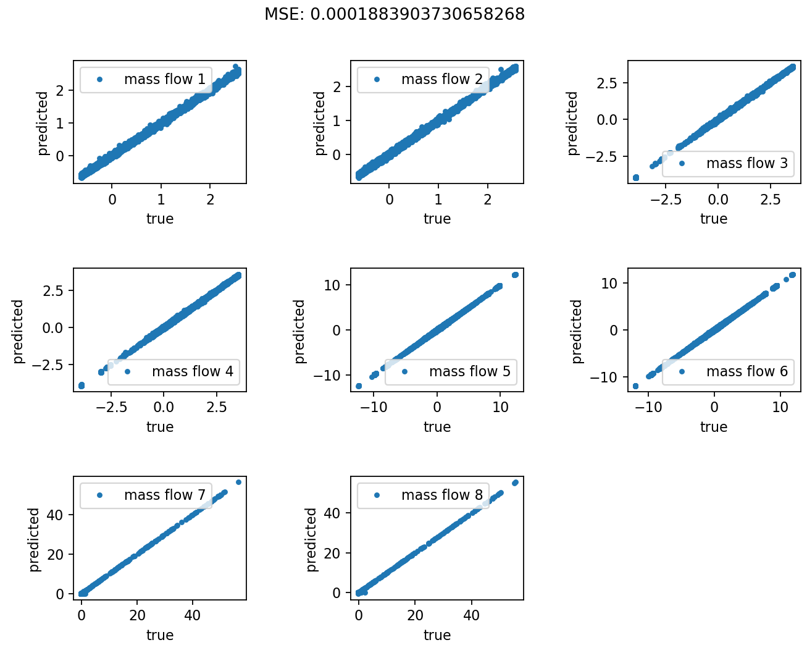}
  \caption{Surrogate model: training results.}
  \label{fig:training results}
\end{figure}

We used the physics-based model to generate data for training the surrogate model. We generated 567,205 training data samples, where the size of the input and the output is 21 and 8, respectively. The model simulations were done using the FMU \cite{blochwitz2011functional} representation of the physics-based model, integrated into Python code. The input consists of valve and pump control signals, and the outputs are represented by mass flow rates. The inputs were randomly drawn from their domain of definition, using the uniform distribution. Since the surrogate model accepts parameters of the physics-based model as inputs, we can simulate faults. We trained the model using Adam algorithm, with a step size 0.001. We used the typical regression models loss function, i.e., the mean square error (MSE) loss function. All other hyper-parameters were left at their default values. The training results are shown in Figure \ref{fig:training results}, where we compared the prediction of the mass flow rates against the ground-truth values. Perfect predictions correspond to  straight diagonal lines. The loss function was minimized at MSE=0.0001.

\begin{figure}[htp!]
  \centering
  \includegraphics[width=0.6\linewidth]{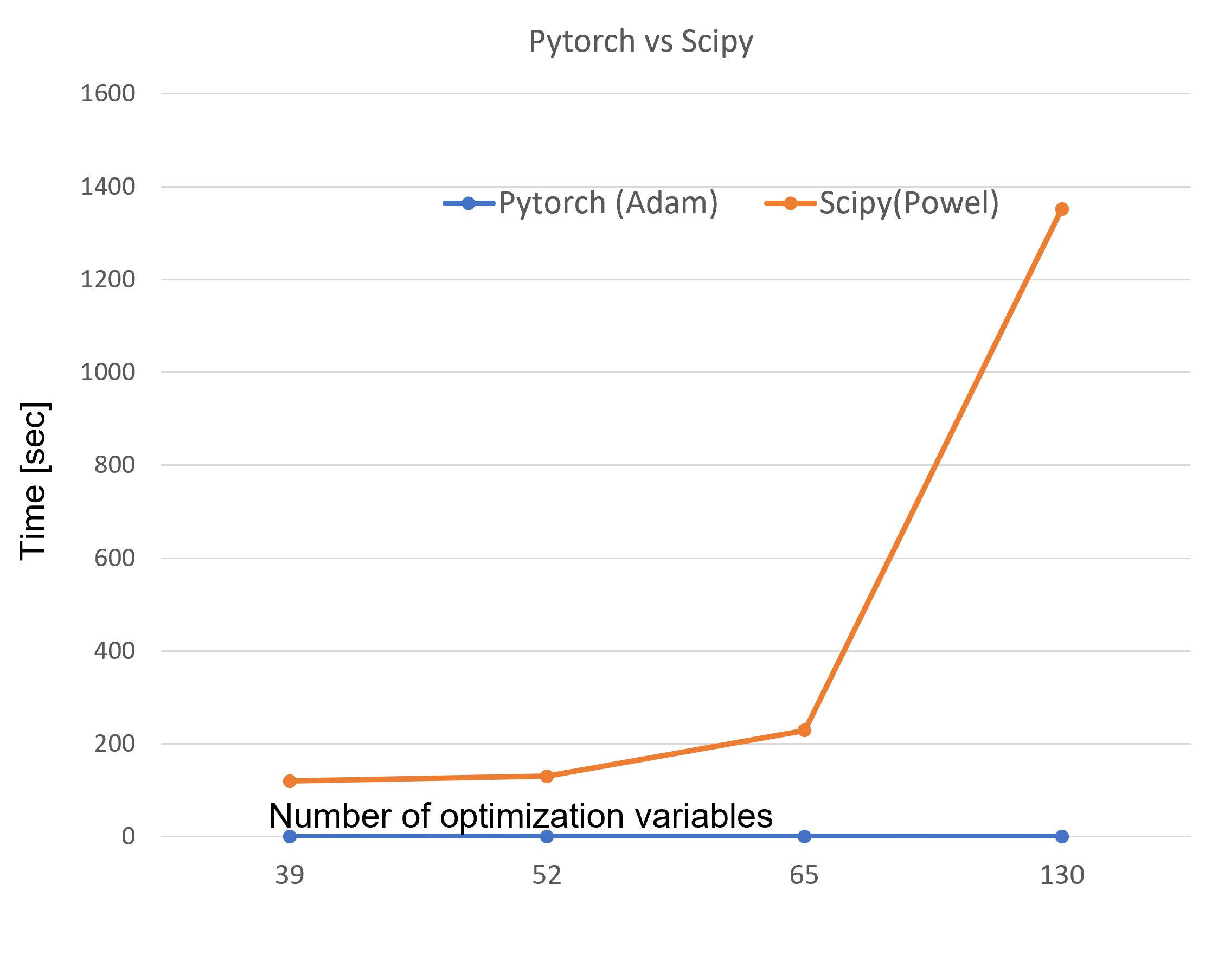}
  \caption{Time to generate a solution for (\ref{eq:10192025}): Pytorch Adam versus Scipy Powell. }
  \label{fig:comparison}
\end{figure}

\begin{figure}[htp!]
  \centering
  \includegraphics[width=0.8\linewidth]{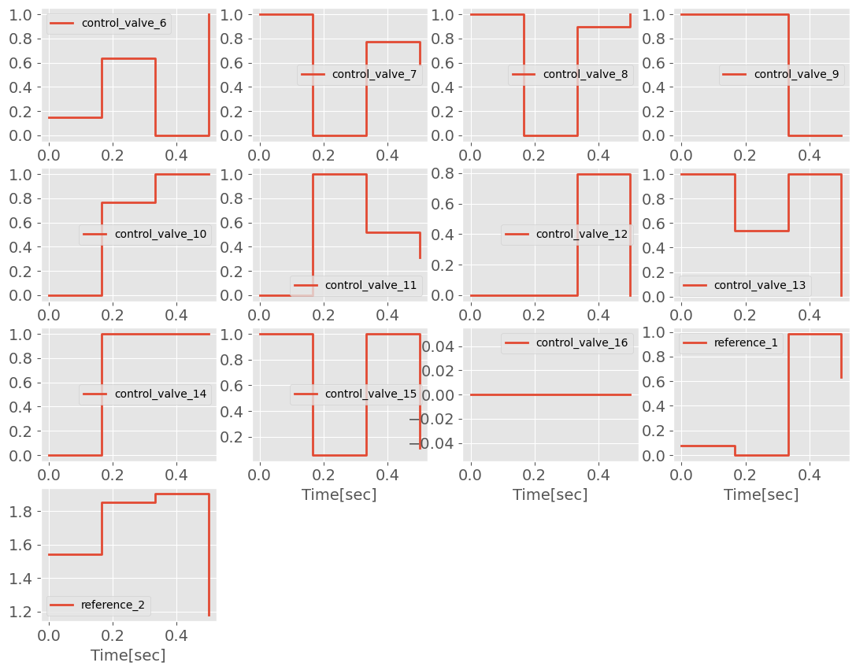}
  \caption{Disambiguation actions generated by the Pytorch implementation of (\ref{eq:10192025}).}
  \label{fig:interventions}
\end{figure}

The goal is to compute actions that yield measurements that distinguish amongst ambiguous diagnoses. We recall that using the 8 mass flow rate sensor, the diagnosis algorithm cannot distinguish between faults 3 and 5, while the diagnosis produced for both faults the correct leak parameter, i.e., 0.75 leak valve opening. We solved the disambiguation optimization problem shown below in Pytorch.
\begin{eqnarray}
\label{eq:10192025}
\min_{\boldsymbol{u}_0, \ldots, \boldsymbol{u}_{\tau}} & & - \sum_{k=0}^{\tau}\|\boldsymbol{\hat{y}}_k^3 - \boldsymbol{\hat{y}}_k^5\|^2\\
\label{eq:eq:10192026}
\textmd{subject to:} &  &\boldsymbol{\hat{y}}_k^i = \hat{h}(\boldsymbol{u}_k; \hat{p}_i), i\in \{3, 5\}, p_i = 0.75,\ \forall i\\
\label{eq:10192027}
 &  & \boldsymbol{u}_k\in \mathcal{U}, \forall k \in \{0, \ldots, \tau\},
\end{eqnarray}
where $\hat{h}$ is the neural network based model, $\mathcal{U} = [0,1]^{11}\times [-5, 5]^2$ defining the constraint set for the valve and pump inputs. Such bound constraints were implemented in Pytorch through clipping.
We solved the optimization problem (\ref{eq:10192025}) using both the Pytorch-based Adam algorithm and Scipy-based Powell algorithm, for various $\tau$ values. We recorded the results and compared them as shown in Figure \ref{fig:comparison}.
Not unexpectedly, in the cased of the Powell gradient-free algorithm, the time to generate a solution increases exponentially with the number of optimization variables. In the case of the Pytorch implementation, the optimization time scales linearly with the number of optimization variables. The difference in efficiency comes from two reasons: (i) Pytorch implementation used batch execution to evaluate the model at multiple inputs as determined by $\tau$, and (ii) the gradient of the loss function is generated through automatic differentiation. Reason (ii) is enabled by surrogate model representation that includes constructs amenable to automatic differentiation. Figure \ref{fig:interventions} shows the actions generated by the Pytorch implementation of the optimization algorithm (\ref{eq:10192025}), for $\tau=4$, where the actions include valve and pump input signals.
Using these actions, we applied again the diagnosis algorithm to the faulty fuel system resulting in the isolation of the correct faults with probability one. 
In addition, the leak parameters are correctly estimated, as shown in Table \ref{tab:high_leak disambiguation}.

\begin{table}[ht!]
\centering
\resizebox{\columnwidth}{!}{%
\begin{tabular}{|c|c|c|c|c|c|c|c|c|}
\hline
 & \textbf{leak\_fault\_1} & \textbf{leak\_fault\_2} & \textbf{leak\_fault\_3} & \textbf{leak\_fault\_4} & \textbf{leak\_fault\_5} & \textbf{leak\_fault\_6} & \textbf{leak\_fault\_7} & \textbf{leak\_fault\_8}\\
\hline
\textbf{leak\_fault\_1} & 0.75 & 0.48 & 0.63 & 0.6 & 0.57 & 0.39 & 0.41 & 0.3\\
\hline
\textbf{leak\_fault\_2} & 0.49 & 0.75 & 0.67 & 0.68 & 0.65 & 0.68 & 1.0 & 0.67\\
\hline
\textbf{leak\_fault\_3} & 0.68 & 0.72 & 0.75 & 0.75 & 0.72 & 0.73 & 1.0 & 0.72\\
\hline
\textbf{leak\_fault\_4} & 0.63 & 0.72 & 0.73 & 0.75 & 0.72 & 0.73 & 1.0 & 0.72\\
\hline
\textbf{leak\_fault\_5} & 0.63 & 0.71 & 0.73 & 0.75 & 0.75 & 0.73 & 1.0 & 0.72\\
\hline
\textbf{leak\_fault\_6} & 0.36 & 0.71 & 0.71 & 0.73 & 0.7 & 0.75 & 1.0 & 0.75\\
\hline
\textbf{leak\_fault\_7} & 0.08 & 0.3 & 0.31 & 0.33 & 0.28 & 0.35 & 0.75 & 0.5\\
\hline
\textbf{leak\_fault\_8} & 0.34 & 0.63 & 0.63 & 0.64 & 0.61 & 0.67 & 1.0 & 0.75\\
\hline
\end{tabular}
}
\caption{Leak fault parameter estimation for the \emph{high} leak case under the disambiguation interventions. For a correct estimation, the diagonal values should correspond to 0.75.}
\label{tab:high_leak disambiguation}
\end{table}
Surrogate models that include both exogenous inputs and parameters can be used for diagnosis as well. One challenge with using surrogate models is that it requires re-training the model if we extend the set of faults we would like to diagnose. In addition, since we test for each fault in parallel using the physics-based model, the gain in time efficiency is not be significant.

\section{Prognostics}
\label{prognostics}

The prognostics module is responsible for (1) constructing a
progression/degradation model of diagnosed faults, and (2) estimating
the remaining useful life (RUL) of the faulted component.\todo{why}

Our approach to prognostics is hybrid: we combine a physics-based model of the system with a data-driven model of degradation. The data-driven degradation model uses diagnosis results to update its parameters after each diagnosis. Extrapolations of the degradation model combined with simulations of the system model enable the assessment of future health of the components and the evaluation of system reliability metrics (e.g., whether or not the mission objectives are achieved).

We model the degradation of a component using a dynamical model for parameter evolution that is trained online. In particular, for each fault parameter $p_i$ we represent its changes over time using Non-linear Autoregressive Models with Moving Average and Exogenous Input (NARMAX) \cite{billings2013nonlinear}:
\begin{equation}\label{eq:11041530}
p_{k} = g(p_{k-1}, \ldots, p_{k-n_p}, e_{k-1}, \ldots, e_{k-n_e}) + e_k,
\end{equation}
where $p_k$ and $e_k$ are the fault parameter and the noise.  The noise represents  the uncertainties in the model. The positive integer $n_p$ and $n_e$ are the maximum delays for the fault parameter and noise, respectively. There are several nonlinear function representations for $g$: neural networks, fuzzy logic-based models, radial basis functions, wavelet basis, polynomial basis or generalized additive model. In this work we represent the map $g$ using polynomial basis, i.e.:
\cite{billings2013nonlinear}:
\begin{equation}\label{eq:11041545}
p_{k} = \Theta_0 + \sum_{i}\Theta_p^ip_{k-1} + \sum_{j}\Theta_e^je_{k-j}+\sum_{i,j}\Theta_{pe}^{ij}p_{k-i}e_{k-j}\\
+ \sum_{i,j}\Theta_{p^2}^{ij}p_{k-i}p_{k-j} + \ldots,
\end{equation}
where $\Theta_0$, $\Theta_p^i$, $\Theta_e^j$, $\Theta_{pe}^{ij}$ and $\Theta_{p^2}^{ij}$ are constant parameters that have to be learned. Depending how many basis terms we choose in the representation, NARMAX models can become fairly complex. Yet, NARMAX models are popular because they can  represent complex systems with sparse nonlinear representations. They select the basis terms that best explain the observations using robust algorithms for model structure selection.

\begin{figure}[htp!]
  \centering
  \includegraphics[width=\linewidth]{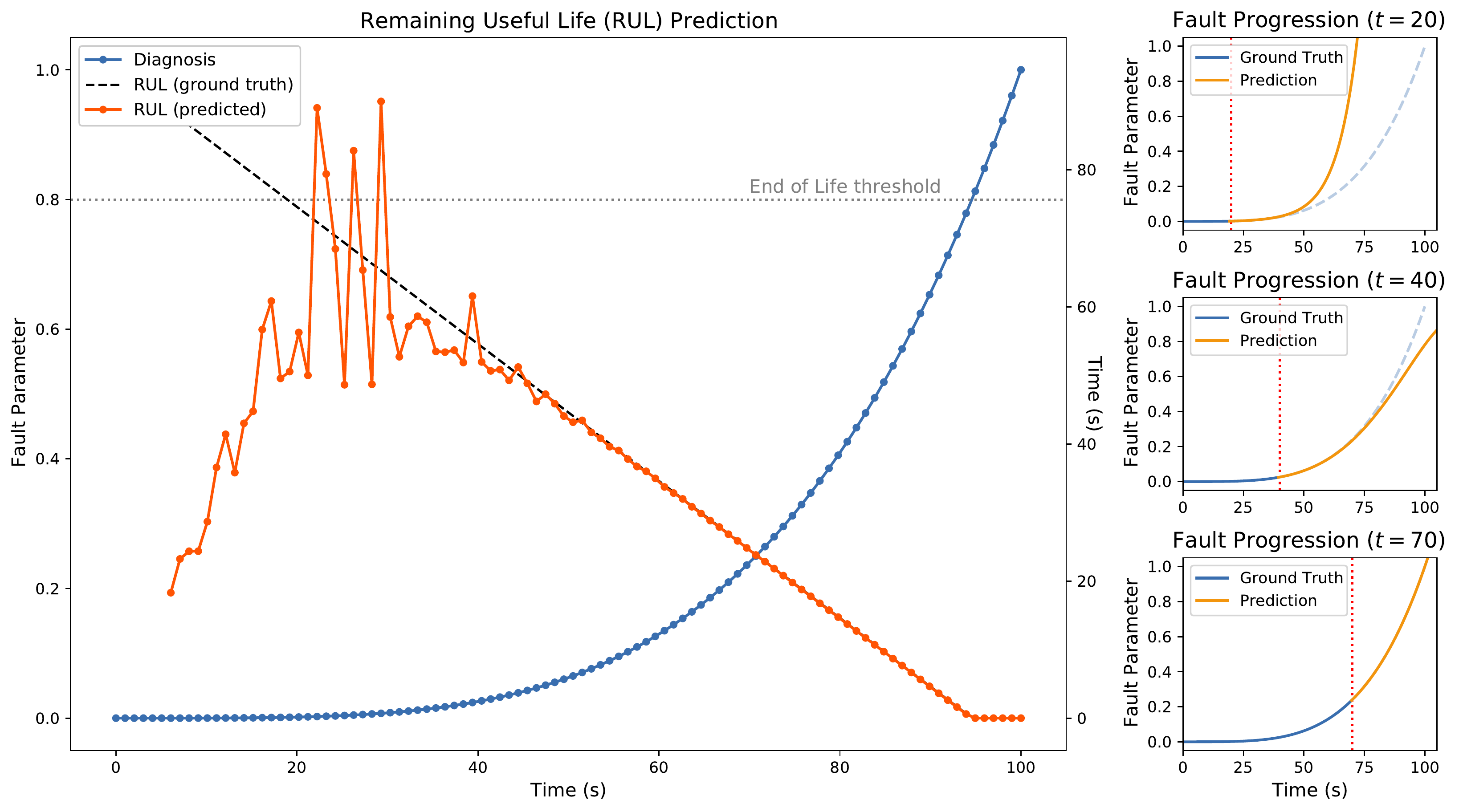}
  \caption{Example prognosis scenario.}
  \label{fig:prognosis_plot}
\end{figure}

Figure~\ref{fig:prognosis_plot} depicts an example prognosis scenario using a simulated
fault progression. In blue we show example diagnoses generated
every second over a 100 second period. Such diagnoses are ingested by the
prognostics module every time the diagnosis module finishes its computation,
prompting it to update its fault progression model. The plots to the right of
Figure~\ref{fig:prognosis_plot} show the computed fault progression prediction
when the prognostics module has diagnostic information of the first 20 seconds,
40 seconds, and 70 seconds, respectively. With more data from diagnostics, our
NARMAX model is able to more accurately track the actual fault progression.

Alongside diagnoses, the prognostics module also takes as input a description of
end of life for every possible fault. This description can either be some fixed
threshold for the faulted parameter, or a boolean function encoding the
relationship between simulation outputs and mission objectives. In the example
depicted in Figure~\ref{fig:prognosis_plot}, we have set an end-of-life
threshold for the faulted parameter of 0.8. Coupled with our degradation model,
we can then determine the remaining useful life (RUL) of the faulted component
by computing when the threshold will be reached (or, given a function encoding
mission objectives, when the system can no longer achieve such objectives). We
show the RUL estimates in red. As the fault magnitude increases, our RUL
estimates more closely track the ground truth, depicted in the plot as a black
dashed line.

\section{Reconfiguration}
\label{sec:reconfiguration}

The reconfiguration module is tasked with adjusting the settings of a system to mitigate the detrimental effects of present faults. Defining the system, its structure and dynamics, requires an expressive language that encompasses a wide range of features, functions, and behaviors. We cast reconfiguration as an automated planning problem and define it in PDDL+~\cite{fox2006modelling}, an extension of the standardized planning modeling language~\cite{mcdermott1998pddl} designed for hybrid systems. Most realistic scenarios, especially in the domain of cyber-physical systems, are inherently hybrid, i.e., exhibiting both discrete mode switches, as well as continuous quantities. PDDL+ planning enables accurate modeling of complex hybrid systems, as well as various additional constraints, goals, and characteristics. Indeed, a realistic planning model is composed of multiple layers of information that require reasoning with to obtain a valid and adequate solution. Our system merges the data from its diagnosis and prognosis modules, and relevant information about the structure and function of the target system, into a hollistic planning model that faithfully reflects the real-world scenario. 

\begin{figure}[hb!]
  \centering
  \includegraphics[width=0.9\linewidth]{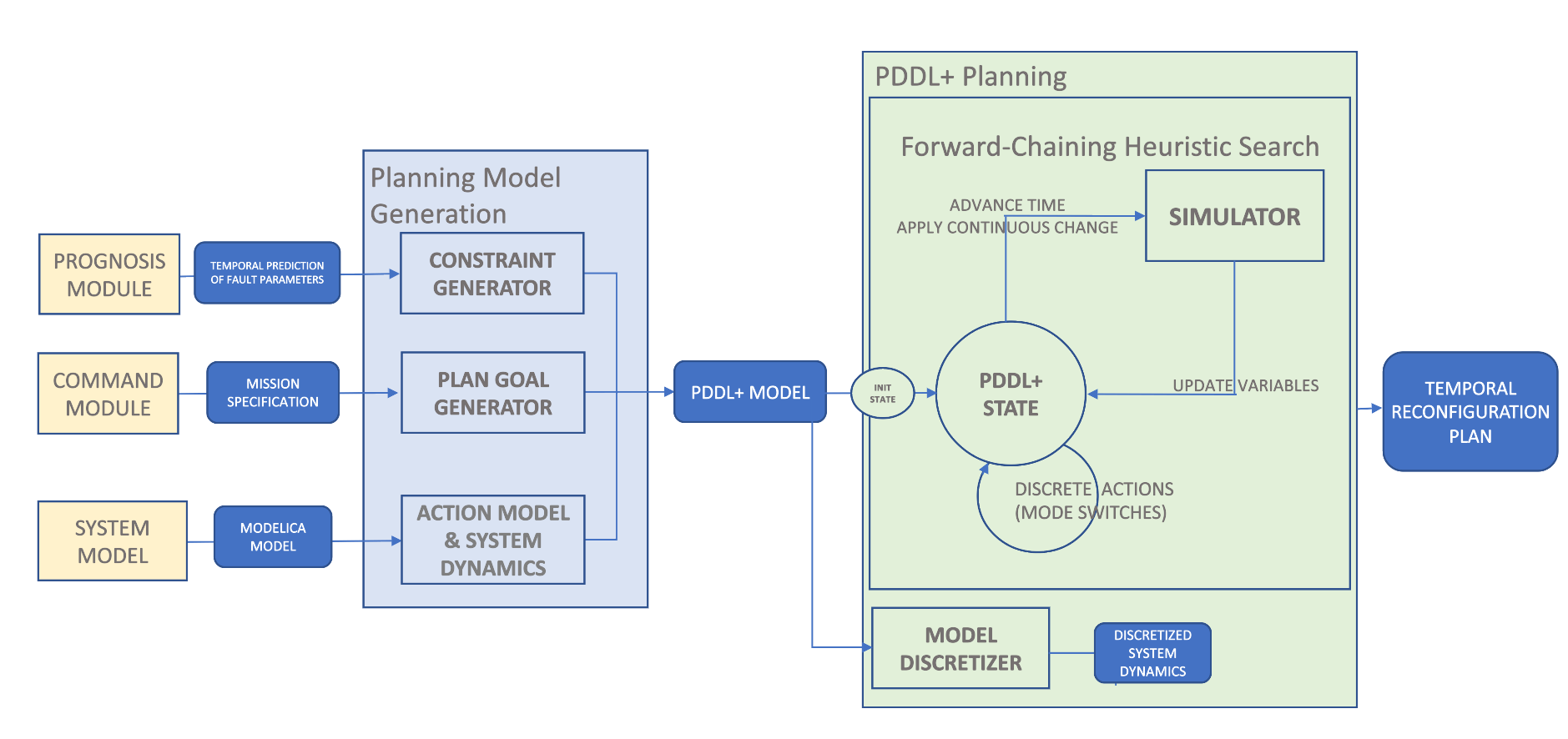}
  \caption{Overview of the Reconfiguration Module}
  \label{fig:reconf_overview}
\end{figure}

A high-level overview of the Reconfiguration approach is depicted in Figure~\ref{fig:reconf_overview}. The PDDL+ model contains the information passed on from other modules. Diagnosis and prognosis modules provide information on the state and temporal evolution of the present fault. This information is encoded into the PDDL+ domain as temporal activity (i.e., processes and events) that update fault information according to the prognosis predictions. 

Planning goal conditions obtained from the mission specification are expressed as planning goal conditions and global problem constraints that any valid solution needs to satisfy. A valid solution to the planning problem is a plan that ends in a state where all goal conditions are satisfied. In the fuel system scenario, the goal condition is reached after the system has been continuously operational for a given duration. 

Constraints are normally enforced by PDDL+ events which are automatically triggered once a constraint is violated (``must-happen'' behavior). Effects of such events irreversibly falsify one or more of the goal conditions, forcing the planner into exploring alternative solutions. In the fuel system's case, examples of constraints are maintaining a sufficient mass flow rate to feed fuel to the engines (Figure~\ref{fig:engine_fuel_constraint}), and not exceeding fuel reserves limited by tank capacity. 

\begin{figure}[!ht]
\begin{center}
\begingroup
    \fontsize{9pt}{10pt}\selectfont
\begin{verbatim}
    (:event engine_massflow_drop
     :parameters (?mfs - massflow_sensor)
     :precondition (and (nominal_status) (engine_massflow ?mfs)
       (or
         (>= (massflow_rate ?mfs) (* (reference_massflow_rate ?mfs) (+ 1 (- 1 (error_margin))) ) )
         (<= (massflow_rate ?mfs) (* (reference_massflow_rate ?mfs) (error_margin)))
       )
     )
     :effect (and (not (nominal_status)))
    )
\end{verbatim}
\endgroup
\caption{Example PDDL+ event defining a constraint which requires that the engines be continuously supplied with sufficient fuel compared to nominal flow.}
\label{fig:engine_fuel_constraint}
\end{center}
\end{figure}

Finally, the PDDL+ domain's system dynamics and action model is defined based on directly on the structure of the target system model (e.g., defined in Modelica). Actions are defined for each reconfigurable component of the system, along with accompanying exogenous activity constructs (i.e., discrete events and continuous processes that are triggered or impacted by executed actions). 


The resulting composable PDDL+ model is then passed into our automated PDDL+ planner. Based on the composition and dynamics of the PDDL+ model, the planner decides if and when to adjust the reconfigurable components of the system to complete the mission objectives while avoiding violating the system's constraints.

\smallskip

Expressive planning models ensure accurate representation of complex systems, though the resulting planning problems are notoriously challenging for planners, particularly severe state-space explosion. To tackle such complex problems, we used \emph{Nyx}, a novel Python-based lightweight PDDL+ planner. Nyx employs a forward-chaining heuristic search to traverse the state space in pursuit of a valid plan (sequence of temporal actions that achieves the reconfiguration objectives). 

Nyx mitigates some of the challenges of hybrid planning by exploiting a \textit{planning-via-discretization} approach where the continuous dynamics of a planning model are approximated using a uniform time-step and step-functions. This requires a fine balance between finding a discretization quantum fine enough to enable finding meaningful solutions, and coarse enough to maintain reasonable efficiency of the search. 

The Nyx planner exploits an anytime search algorithm. Instead of terminating after finding the first-encountered feasible plan (as is the norm in AI planning), Nyx will return all solutions that were found within a given run-time budget. 
The obtained set of solutions are ranked according to any relevant metric (e.g., minimize leaked fuel, maximize fuel delivered to the engines, minimize number of valve and pump switches). 

Many systems are governed by complex dynamics beyond the scope of what can be expressed even with PDDL+. To circumvent this limitation, the Nyx planner is equipped with semantic attachments~\cite{dornhege2009semantic}, i.e., external functions to which part or all of the model's continuous dynamics are expressed. Often, the semantic attachment is in the form of a learned surrogate model. In the case of the fuel system, we delegate the fluid dynamics calculations to an external function in the form of an attached FMU. We pass the planning state (containing the valve and pump settings) as inputs to the FMU and then simulate it over the duration of the planner's discretized time-step. After the simulation, the extracted values are passed back to the planner and used to update the values of the state variables\footnote{For the sake of clarity, we highlight that the information on the faulty component and its evolution can be explicitly encoded in the PDDL+ domain as a state variable with associated happenings (e.g. a continuous degradation process), where the variable is one of the inputs to the attached FMU. On the other hand, fault information can also be obscured from the planner inside the FMU (i.e., making the planner oblivious to the fault model, only reasoning with the fault's impact on the FMU-updated state variables).}.
During the search phase, whenever a continuous change takes place in the system, a simulation is performed using the external semantic attachment, all other changes are assumed to be discrete (i.e., mode switches). The search cycle as well as the external function calls are outlined in the \emph{PDDL+ Planning} box in the diagram depiction of the Reconfiguration Module (Fig.~\ref{fig:reconf_overview}). 



\begin{table}[]
\resizebox{\textwidth}{!}{%
\begin{tabular}{|l|l|l|}
\hline
Happening &
  Description &
  Fuel System Examples \\ \hline
Actions &
  \begin{tabular}[c]{@{}l@{}}Direct executive, the planner can choose to apply when \\ preconditions are satisfied.\end{tabular} &
  \begin{tabular}[c]{@{}l@{}}- Valve Control: increase/decrease the position of a valve by a specified interval. \\   The interval of 1 means that the valve can only be fully open or fully closed.\\ - Pump Control: increase/decrease the pump rotational speed by a given interval.\end{tabular} \\ \hline
Events &
  \begin{tabular}[c]{@{}l@{}}A set of exogenous happenings with discrete effects which must \\ be applied immediately when their preconditions are satisfied.\end{tabular} &
  \begin{tabular}[c]{@{}l@{}}- Fuel massflow rate constraints.\\ - Fuel usage constraints.\end{tabular} \\ \hline
Processes &
  \begin{tabular}[c]{@{}l@{}}A set of exogenous happenings with continuous effects which \\ must be applied for as long as the processes' conditions hold \\ true. The agent cannot directly influence processes.\end{tabular} &
  \begin{tabular}[c]{@{}l@{}}- Leak fault degradation.\\ - Tracking cumulative fuel usage.\\ - Time-keeping.\end{tabular} \\ \hline
State Variables &
  Fields that define the current state of the system. &
  \begin{tabular}[c]{@{}l@{}}- Position of valves and pumps.\\ - Massflow rate sensor values and reference points.\\ - Active leaks and their intensity.\end{tabular} \\ \hline
Auxiliary Variables &
  Additional variables used to accurately define the fuel system. &
  \begin{tabular}[c]{@{}l@{}}- Bounds on valve positions and pump settings.\\ - Valve and pump adjustment intervals.\\ - Cumulative fuel usage values (spent, fed to engines, nominal values).\\ - Time-keeping (elapsed time, delay, time limits, etc).\end{tabular} \\ \hline
\end{tabular}%
}
\caption{Composition of the PDDL+ planning model for the fuel system scenario.}
\label{tab:pddl_fuel_system_composition_}
\end{table}

\medskip

A solution to the fuel system reconfiguration is a sequence of timed actions which adjust the positions of valves and pump speeds, such that the system is operational for a given duration without violating any of it's constraints. The plan specifies when to change the positions and settings of specific reconfigurable components such that mission objectives are achieved. 


\subsection{Experimental Results}
To highlight the flexibility and applicability of our planning approach,
we present results for the reconfiguration of the fuel system in two double fault scenarios.  In the first scenario, we consider a valve stuck open and a leaky pipe (that is static in intensity), where the mission requires immediate isolation of the fault and limiting the fuel loss. The second scenario is more complex: a valve stuck closed and a leaky pipe which worsens over time. They are set up in such a fashion that not using the faulty components would prevent achieving mission objectives, thus the reconfiguration module must reason with and use the faulted components.

\begin{table}[]
\caption{Planning problem setup common across both scenarios. }
\label{tab:planning_problem_setup}
\resizebox{\textwidth}{!}{%
\begin{tabular}{|l|l|}
\hline
Type & Description                                                                                       \\ \hline
Goal & deliver 30 gallons of fuel per hour to each engine for a continuous period of 10 hours.           \\ \hline
Actions &
  \begin{tabular}[c]{@{}l@{}}- adjust the opening of valves (11 count). Opening interval 1 - open/close only.\\ - increase/decrease speed of pumps (2 count). Adjustment interval +/-- 1.\end{tabular} \\ \hline
Time & \begin{tabular}[c]{@{}l@{}}- discretized time step = 1 hour\\ - time horizon = 10 hours\end{tabular} \\ \hline
Constraints &
  \begin{tabular}[c]{@{}l@{}}- massflow at engine \textless  130\% of nominal reference point\\ - massflow at engine \textgreater  70\% of nominal reference point\end{tabular} \\ \hline
Initial State &
  \begin{tabular}[c]{@{}l@{}}
- Cross-valves closed (valves 8, 15, 16)\\
- All other valves open (valves 6, 7, and 9-14)\\
- Nominal pump speed = 3.035\\
- Allowed pump speed range [0, 5]
\end{tabular} \\ \hline
\end{tabular}%
}
\end{table}

\subsubsection{Scenario 1: Static Leak}

\begin{figure}[htp!]
  \centering
  \includegraphics[width=0.99\linewidth]{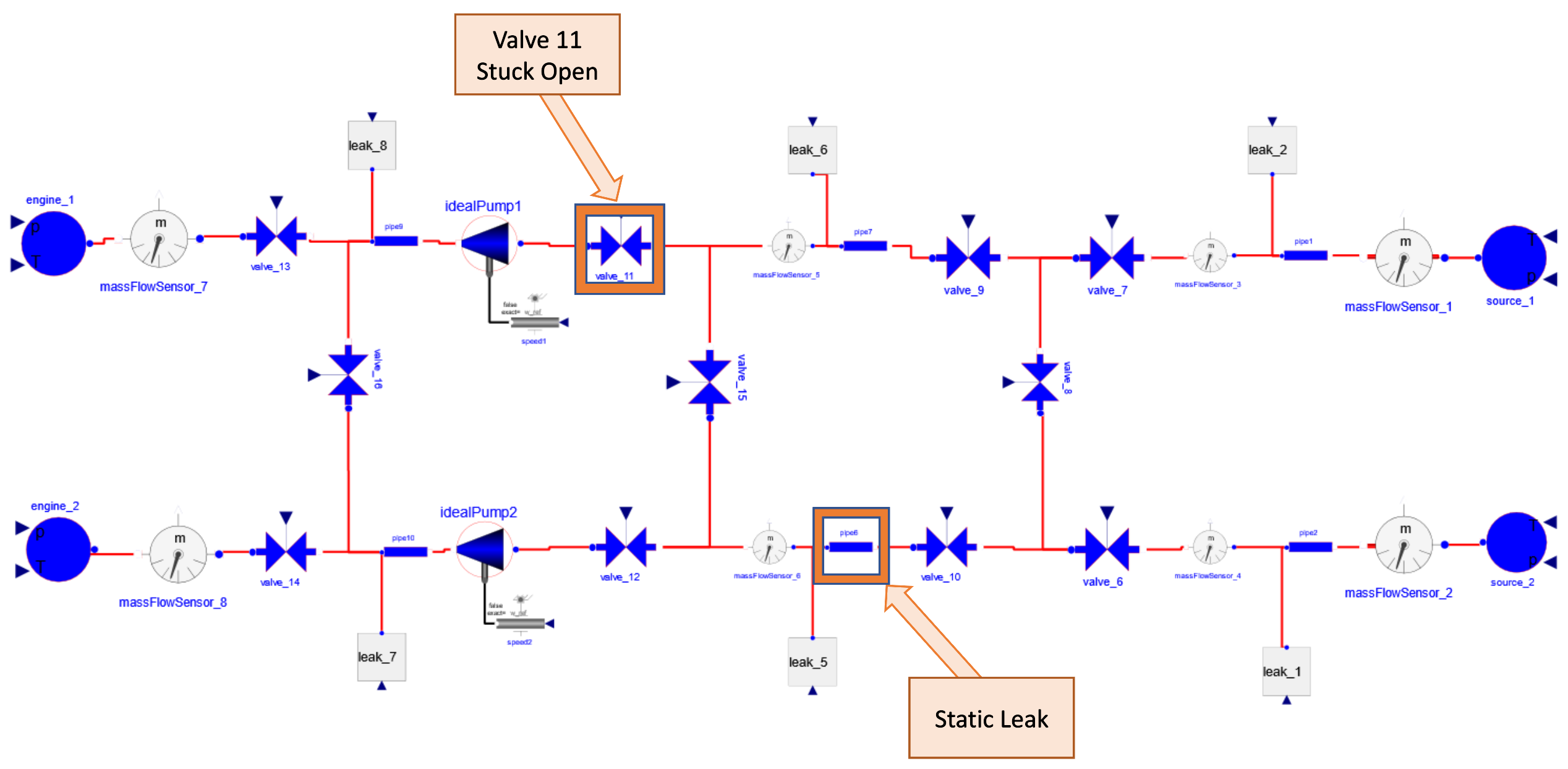}
  \caption{Fault Scenario 1 Setup.}
  \label{fig:reconf_exp_1}
\end{figure}

The first scenario is depicted in Figure~\ref{fig:reconf_exp_1} and includes two faults: valve 11 stuck open, and a leaky pipe 6 (with constant intensity of 0.7). This scenario requires immediately isolating the leak, forcing the planner to reconfigure the system at the beginning of the mission to minimize fuel loss. The fuel tanks hold a total of 630 gallons, which is just enough to run both engines at 30 Gal/hr for 10 hours (with an additional 30 Gal reserve).

\begin{figure}[htp!]
  \centering
  \includegraphics[width=0.99\linewidth]{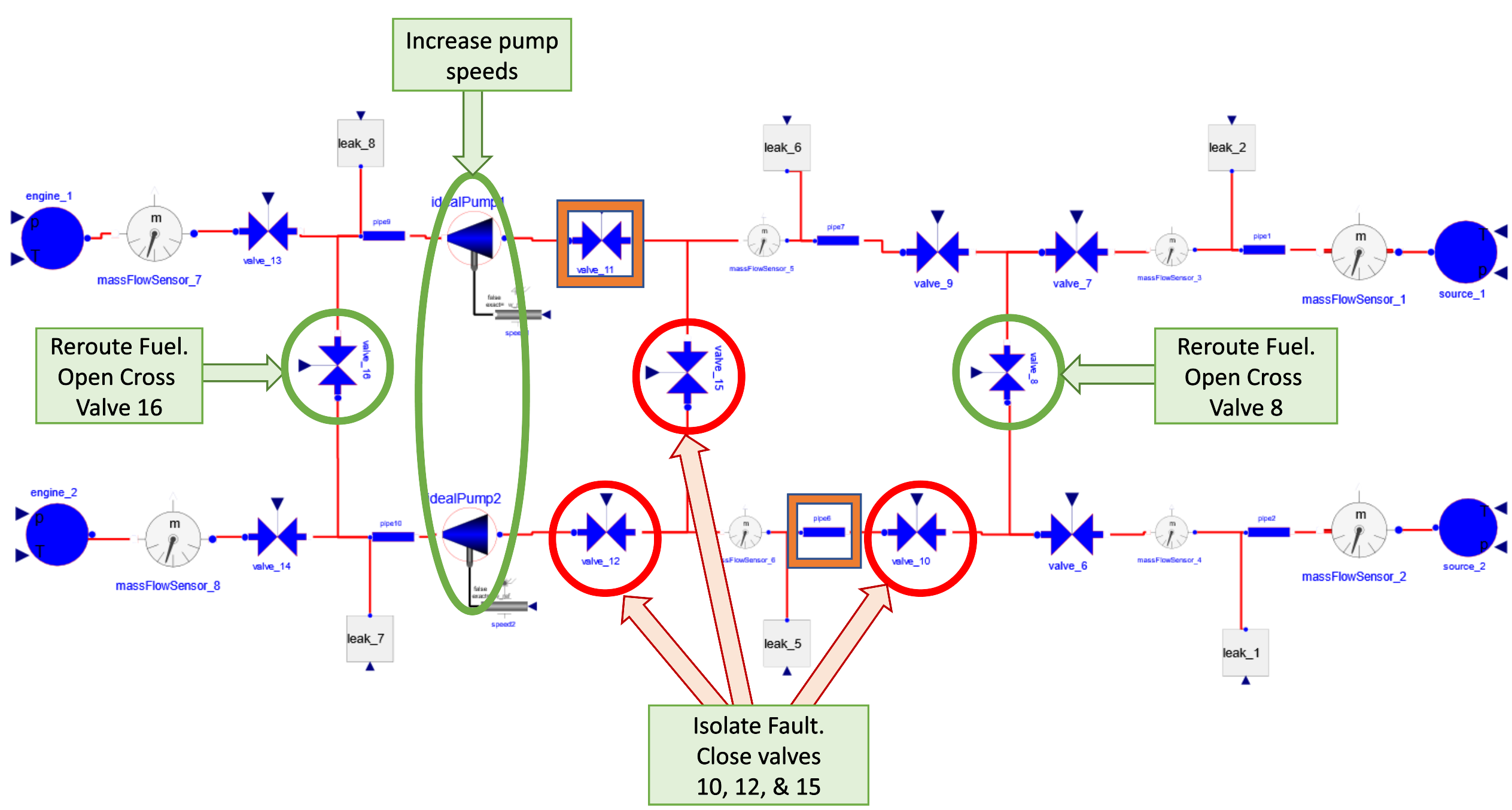}
  \caption{Scenario 1: reconfigured system (green circles represent valves that were opened due to reconfiguration, while the red circles represent valves that were closed. Orange rectangles denote faulted components).}
  \label{fig:reconf_exp_1_solution_diag}
\end{figure}

\begin{figure}[!ht]
\begin{center}
\begingroup
    \fontsize{9pt}{10pt}\selectfont
\begin{verbatim}
                    0.000: open valve_8         [0.0]
                    0.000: open valve_16        [0.0]
                    0.000: close valve_10       [0.0]
                    0.000: close valve_12       [0.0]
                    0.000: close valve_15       [0.0]
                    0.000: increase_pump pump_1 [0.0]
                    0.000: increase_pump pump_2 [0.0]
\end{verbatim}
\endgroup
\caption{Reconfiguration plan for the Fuel System in Scenario 1 (duration=10 hours). The initial numbers specify timing of the actions (i.e., 0.000 means action executed at time t = 0.0, the very beginning of the mission). Next is the action name and its parameter, and ending with the duration of the action ([0.0] means instantaneous action).}
\label{fig:scenario_1_plan}
\end{center}
\end{figure}

Figure~\ref{fig:reconf_exp_1_solution_diag} shows a visual representation of the reconfigured system in scenario 1, according to the solution found by the PDDL+ planner (shown in Figure~\ref{fig:scenario_1_plan}). The reconfiguration module isolates the leaky pipe 6 by closing the surrounding valves 10, 12 and 15, followed by rerouting the fuel around the fault by opening cross valve 8 and increasing the speed of the pumps to deliver the fuel at a sufficient rate to both engines. All reconfiguration actions are executed simultaneously as soon as possible, i.e., at time t = 0.0, to preempt any further fuel loss.

\begin{figure}[!htb]
\begin{subfigure}{0.5\textwidth}
\includegraphics[width=0.9\linewidth, height=6cm]{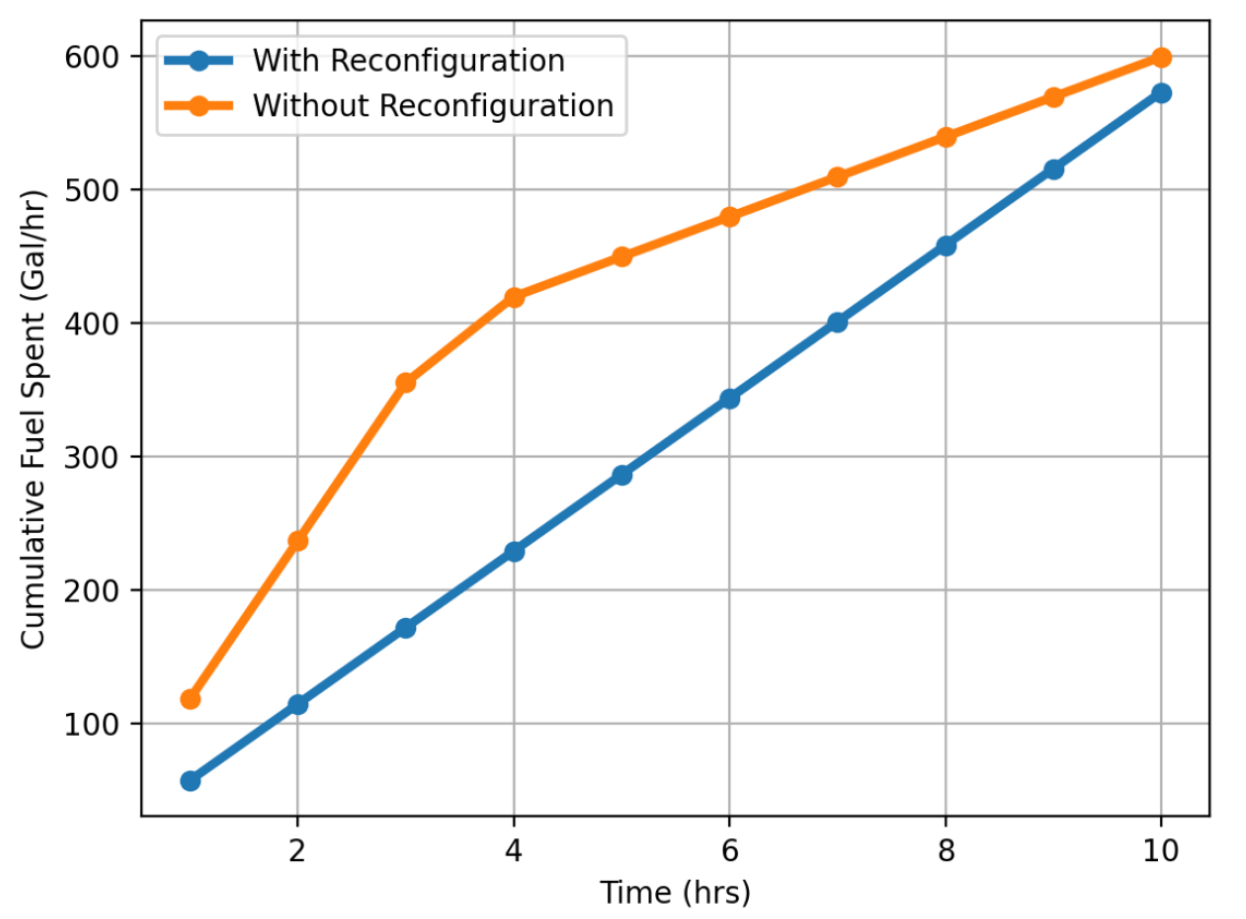}
\caption{Cumulative fuel drawn from the tanks.}
\label{fig:scenario_1_fuel_spent}
\end{subfigure}
\begin{subfigure}{0.5\textwidth}
\includegraphics[width=0.9\linewidth, height=6cm]{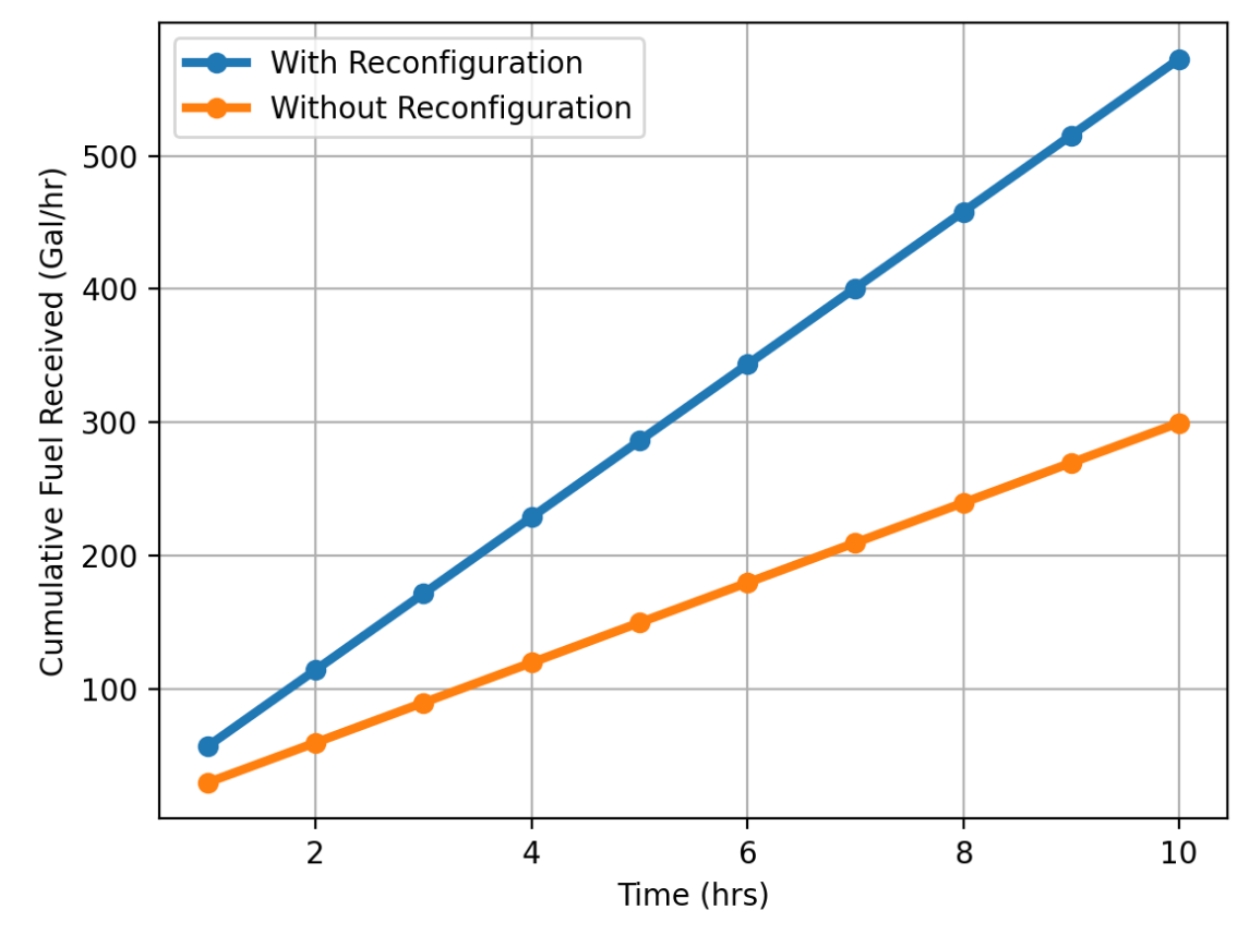}
\caption{Cumulative fuel fed to the engines.}
\label{fig:scenario_2_fuel_received}
\end{subfigure}
\caption{Reconfigured and unchanged system fuel usage for Scenario 1.}
\label{fig:scenario_1_fuel_usage}
\end{figure}

A comparison of the results between the reconfigured system and the unchanged system shows a stark difference, and is described in Table~\ref{tab:scenario_1_fuel_table}. Without reconfiguration, the faulted pipeline fails to deliver any fuel to the engine, leaking 300 Gal of fuel into the vessel in approximately 3.5 hours. On the other hand, the reconfigured system delivers 95.51\% of fuel to the engines compared to an unfaulted system in nominal conditions, while losing only 0.13\% of the spent fuel. The fuel lost in both cases is subject to some rounding errors stemming from the FMU simulation, presumably due to minute amounts of fuel remaining in the pipes after flowing out the fuel tanks. 
The advantages of reconfiguration with respect to fuel usage can be further analyzed by tracking the cumulative fuel siphoned from the tanks and delivered to the engines in Figure \ref{fig:scenario_1_fuel_usage}.

\begin{table}[]
    \centering
    \begin{tabular}{ |c|c|c| }
     \hline
        & Fuel Delivered & Fuel Lost \\
        & (\% of nominal) & \\ \hline
     \textbf{Reconfigured System} & 95.51\% & 0.13\% \\ \hline
     \textbf{Unchanged System} & 49.93\% & 49.96\% \\ \hline
     \hline
    \end{tabular}
    \caption{Scenario 1: Reconfigured system fuel loss comparison against an unchanged system.}
    \label{tab:scenario_1_fuel_table}
\end{table}

\subsubsection{Scenario 2: Worsening Leak}

\begin{figure}[htp!]
  \centering
  \includegraphics[width=0.99\linewidth]{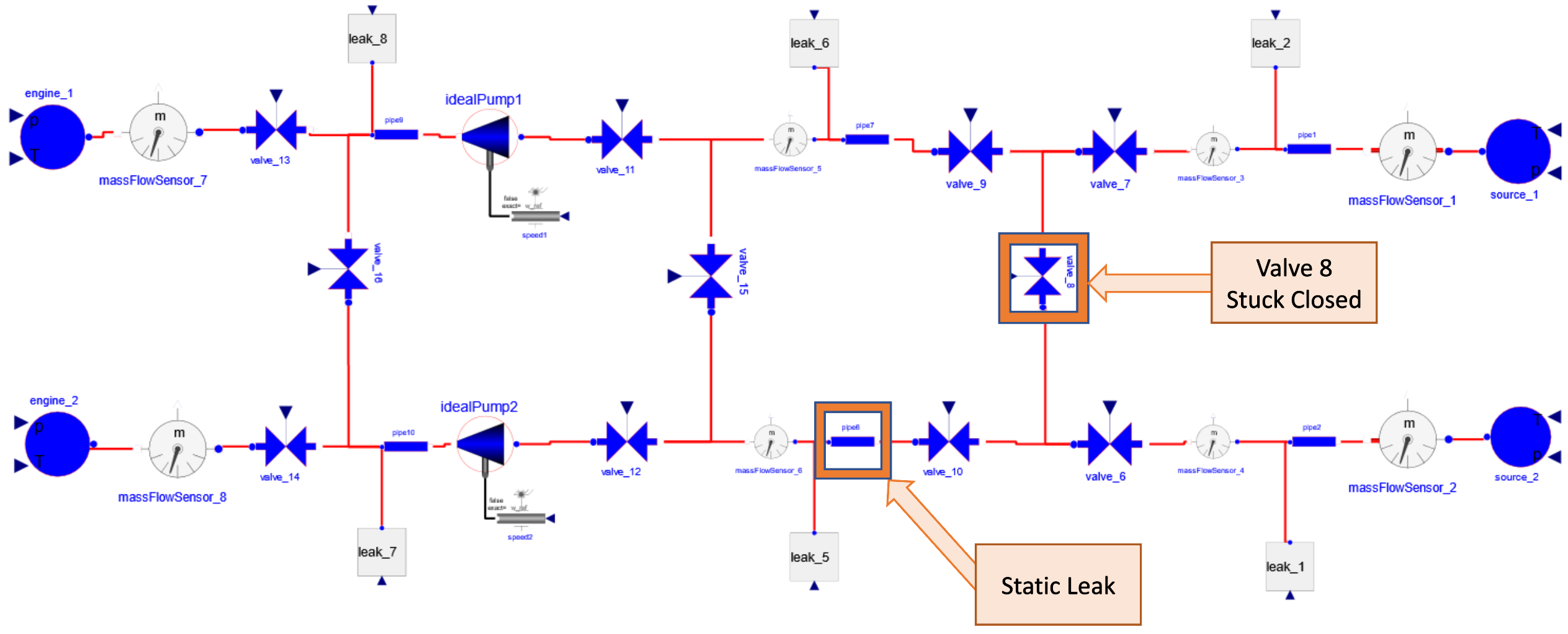}
  \caption{Fault Scenario 2 Setup.}
  \label{fig:reconf_exp_2}
\end{figure}

Scenario 2 of the fuel system is depicted in Figure~\ref{fig:reconf_exp_2} and contains two faults: valve 8 stuck shut, and a leaky pipe 6. The intensity of the leak in the initial state is 0.1 but increases by 0.1 per hour. This configuration of the system and the placement of the faults prevents the planner from isolating the leak and rerouting the fuel around it (as was done in scenario 1). In order to mitigate the leak, the reconfiguration module can shut off one of the fuel tanks. However, doing that too early would result in a failure as one tank does not hold enough fuel to supply both engines simultaneously. The reconfiguration module has no alternative but to allow some fuel to leak into the vessel. In this case, the fuel tanks hold a total of 750 gallons, which is sufficient to supply fuel to both engines at 30 Gal/hr for 10 hours (with a reserve for deliberate leakage).

\begin{figure}[htp!]
  \centering
  \includegraphics[width=0.99\linewidth]{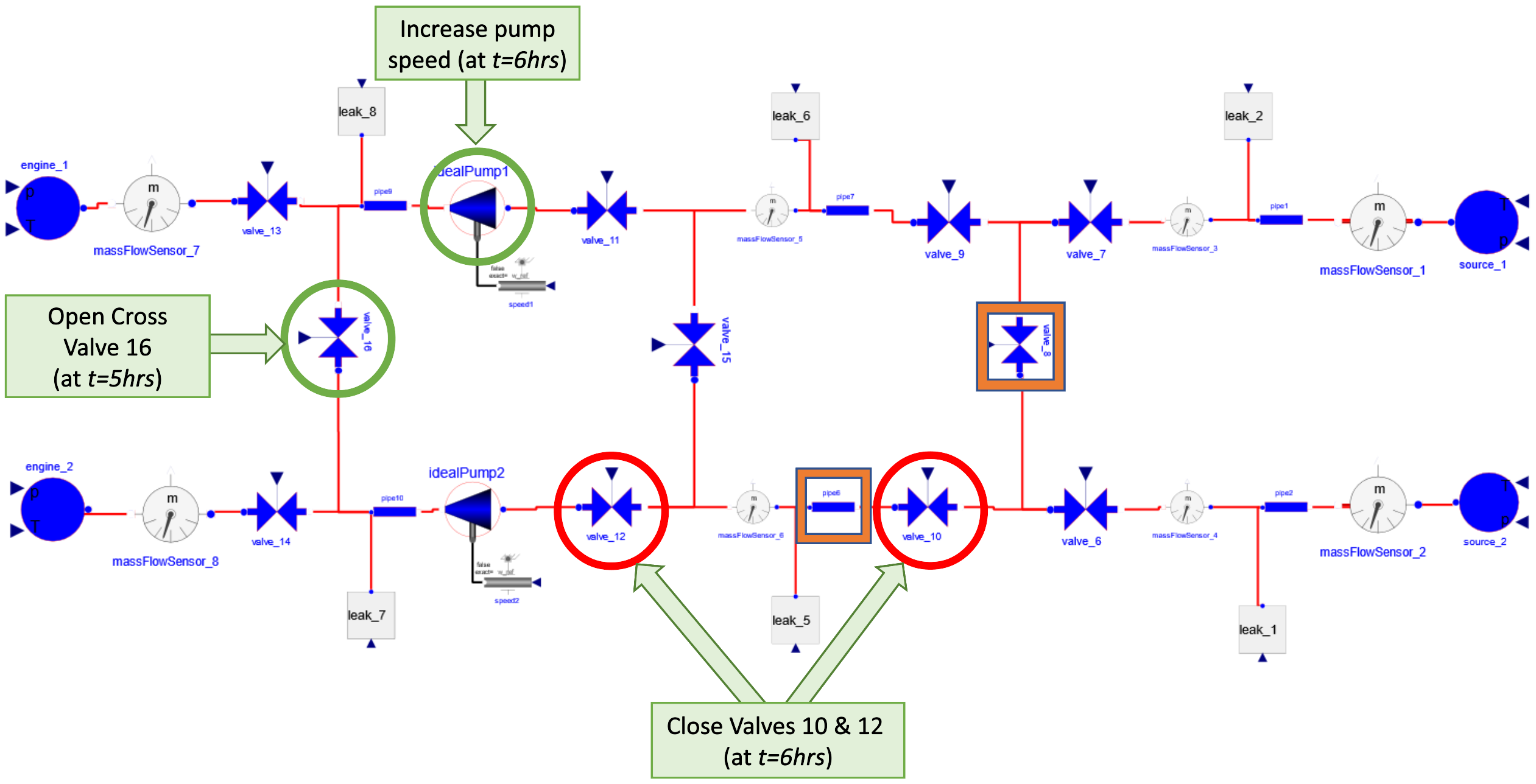}
  \caption{Scenario 2: reconfigured system (green circles represent valves that were opened due to reconfiguration, while the red circles represent valves that were closed. Orange rectangles denote faulted components).}
  \label{fig:reconf_exp_2_solution_diag}
\end{figure}

The reconfigured system for Scenario 2 is depicted in Figure~\ref{fig:reconf_exp_2_solution_diag}, followed by the temporal plan outlined in Figure~\ref{fig:scenario_2_plan}, which shows the adjustments made to valve positions and pump speeds after 5 hours, once the leak in pipe 6 becomes uncontrollable. The reconfiguration is achieved by opening the cross valve 16 at t = 5hr and increasing the speed of pump 1 at t = 6 hr to distribute the fuel from tank 1 between both engines. Simultaneously, at time t = 6hr, valve 10 is closed, shutting off tank 2, and valve 12 is closed to prevent a backwards leak, isolating the leak in the process (since cross valve 15 is closed by default).

\begin{figure}[!htb]
\begin{subfigure}{0.5\textwidth}
\includegraphics[width=0.9\linewidth, height=6cm]{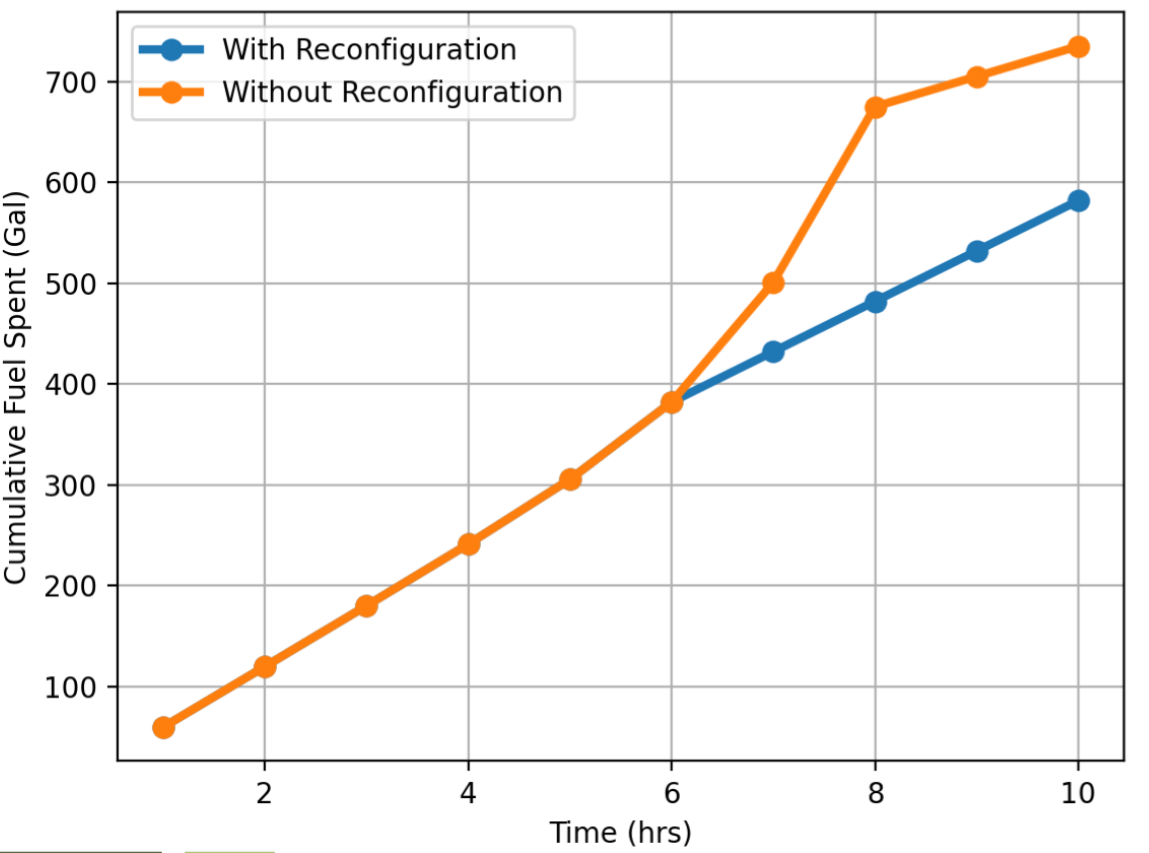}
\caption{Cumulative fuel spent from the tanks.}
\label{fig:scenario_2_fuel_spent}
\end{subfigure}
\begin{subfigure}{0.5\textwidth}
\includegraphics[width=0.9\linewidth, height=6cm]{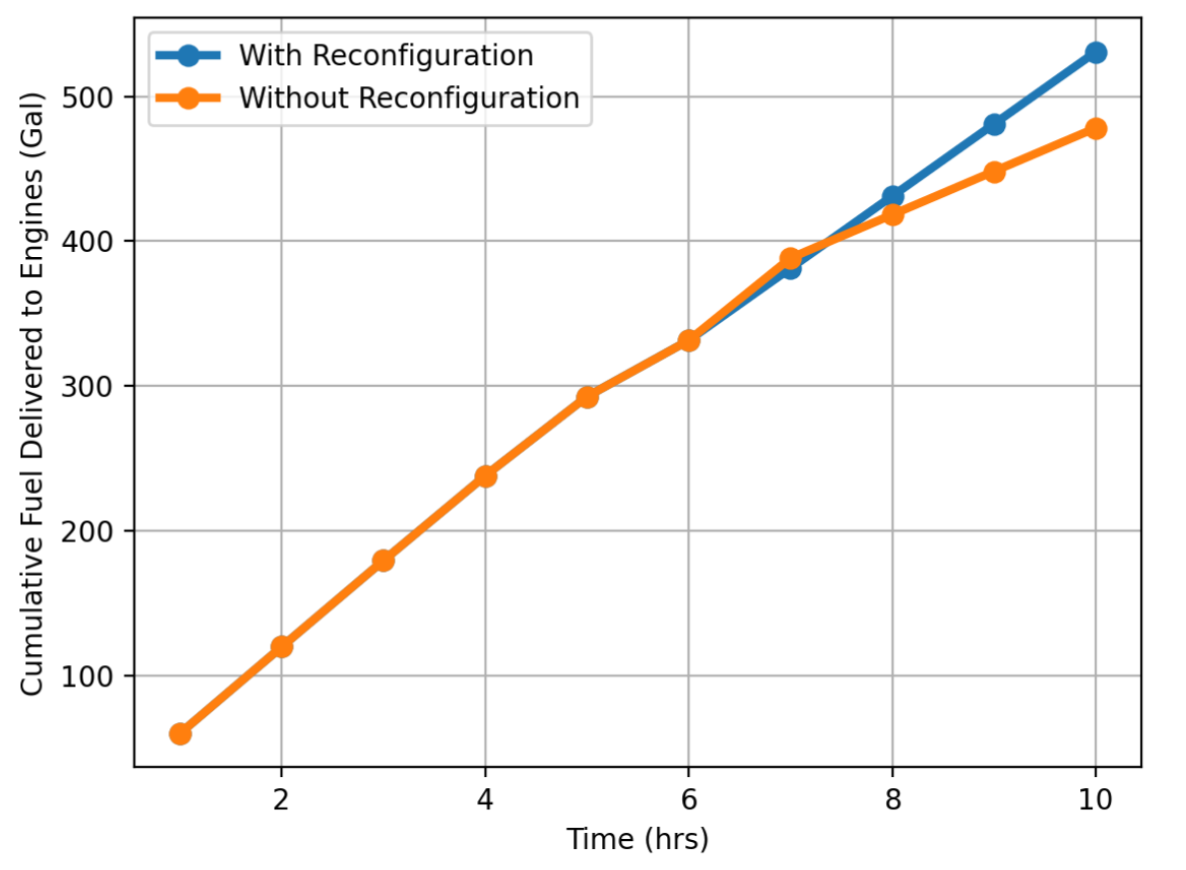}
\caption{Cumulative fuel fed to the engines.}
\label{fig:scenario_2_fuel_received}
\end{subfigure}
\caption{Reconfigured and unchanged system fuel usage for Scenario 2.}
\label{fig:scenario_2_fuel_usage}
\end{figure}

By examining the results from Table~\ref{tab:scenario_2_fuel_table}, it is apparent that the reconfigured system performs well, losing only 3.66\% of the fuel, while delivering 91.39\% of fuel to the engines compared to an unfaulted system in nominal state.
On the other hand, while the unchanged system delivered 82.2\% of cumulative fuel to the engines, compared to the nominal system, it had to spend a substantial overhead of the tank reserves, losing 34.93\% of spent fuel. In fact, in the unchanged system, the leak would cause tank 2 to run dry after 7.55 hours. The differences in fuel usage (spent and received) between the reconfigured and unchanged system are further highlighted in Figure~\ref{fig:scenario_2_fuel_usage}.

\begin{table}[]
    \centering
    \begin{tabular}{ |c|c|c| }
     \hline
        & Fuel Delivered & Fuel Lost \\
        & (\% of nominal) & (spent/received difference) \\ \hline
     \textbf{Reconfigured System} & 91.39\% & 3.66\% \\ \hline
     \textbf{Unchanged System} & 82.2\% & 34.93\% \\ \hline
     \hline
    \end{tabular}
    \caption{Scenario 2: Reconfigured system fuel loss comparison against an unchanged system.}
    \label{tab:scenario_2_fuel_table}
\end{table}

\begin{figure}[!ht]
\begin{center}
\begingroup
    \fontsize{9pt}{10pt}\selectfont
\begin{verbatim}
                    5.000: open valve_16        [0.0]
                    6.000: close valve_10       [0.0]
                    6.000: close valve_12       [0.0]
                    6.000: increase_pump pump_1 [0.0]
\end{verbatim}
\endgroup
\caption{Reconfiguration plan for the Fuel System in Scenario 2 (duration=10 hours).}
\label{fig:scenario_2_plan}
\end{center}
\end{figure}

\section{End-to-end results}
\label{sec:end-to-end-results}
\begin{figure}[htp!]
  \centering
  \includegraphics[width=0.7\linewidth]{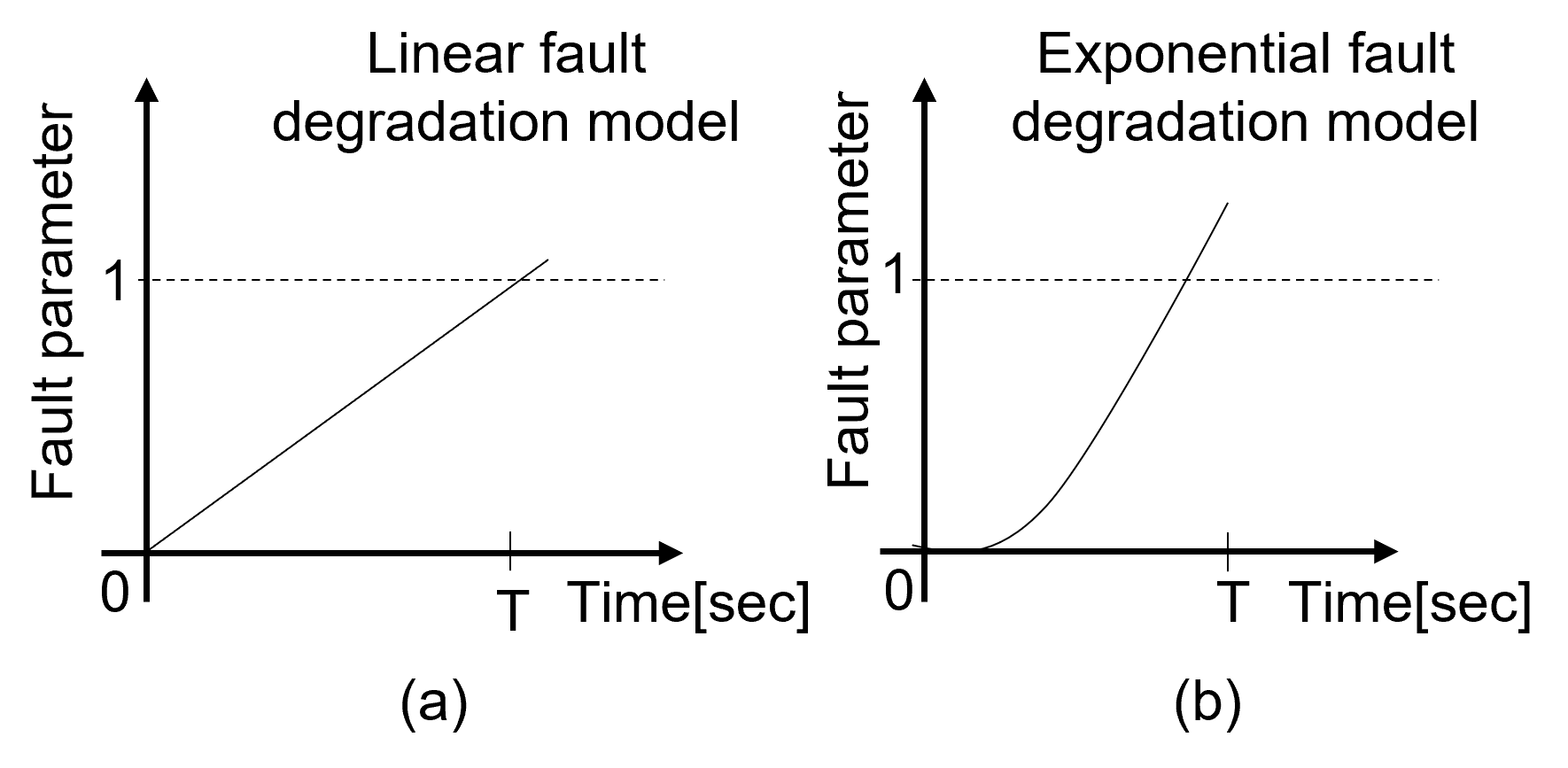}
  \caption{Fault degradation profiles: (a) linear; (b) exponential}
  \label{fig:fault_profiles}
\end{figure}

We evaluate the resilience metric define in (Equation \ref{eq:11050939}) under three scenarios, as shown in Table \ref{tab:resilience scenarios}. These scenarios include leak faults only, and are chosen to demonstrate the impact of the diagnosis, prognostics and reconfiguration modules on satisfying the system requirements. In \textbf{Scenario 1}, we include no health monitoring feature, while in \textbf{Scenario 2} we only include only the diagnosis and the reconfigurations modules. The goal of \textbf{Scenario 2} is to demonstrate the resilience cost we pay when no prognostics is used. In \textbf{Scenario 3} we consider the impact of all three modules.
\begin{table}[ht!]
\centering
\begin{tabular}{|c|c|c|c|}
\hline
 & \textbf{Diagnosis} & \textbf{Prognostics} & \textbf{Reconfiguration}\\
\hline
\textbf{Scenario 1} & \xmark & \xmark & \xmark \\
\hline
\textbf{Scenario 2} & \cmark & \xmark & \cmark \\
\hline
\textbf{Scenario 3} & \cmark & \cmark & \cmark \\
\hline
\end{tabular}
\caption{Scenarios for evaluating the resilience metric}\label{tab:resilience scenarios}
\end{table}
For each scenario, we report the average system resilience over all considered faults. As shown in Figure \ref{fig:fault_profiles}, for each fault we consider two types of degradation profiles: linear and exponential. The fault parameter takes values in [0, 1], with 0 corresponding to no leak, and 1 corresponding to a catastrophic loss of fuel. The RUL of a pipe is defined as the average time until the catastrophic fault occurs, i.e., the fault parameter reaches its upper bound.

\emph{Diagnosis setup}: We consider a time horizon of 1800 seconds, and a diagnosis windows of 10 seconds. In other words we have a total 180 non-overlapping diagnosis windows. During each window, we sample the measurements at a 1 second sampling period, resulting in 10 samples per diagnosis window. For each leak, fault we consider the following linear and exponential degradation models:
$$p(t) = \left\{
\begin{array}{cc}
0 & t<0,\\
\frac{1}{1200}t & t\in [0, 960],\\
0.8 & t>960,
\end{array}
\right.
$$
$$p(t) = \left\{
\begin{array}{cc}
0 & t<500,\\
2^{\frac{t-500}{400}}-1 & t\in [500, 839.2],\\
0.8 & t>839.2,
\end{array}
\right.
$$
According to these models, the linear and exponential degradations reach failure after 960 and 839.2  seconds, respectively. For each scenario we evaluate the resilience metric defined in (\ref{eq:11050939}) using the parameters shown in Table \ref{tab:resilience metric parameters}.
\begin{table}[ht!]
\centering
\begin{tabular}{|c|c|c|c|c|c|c|}
\hline
$y_1^d$ & $y_2^d$ & T & $\beta_1$ & $\beta_2$ & $\alpha_1$ & $\alpha_2$\\
\hline
1 kg/s & 1 kg/s & 1800 s & 3600 kg & 3600 kg & 0.5 & 0.5 \\
\hline
\end{tabular}
\caption{Resilience metric parameters}
\label{tab:resilience metric parameters}
\end{table}

\begin{figure}[htp!]
  \centering
  \includegraphics[width=0.7\linewidth]{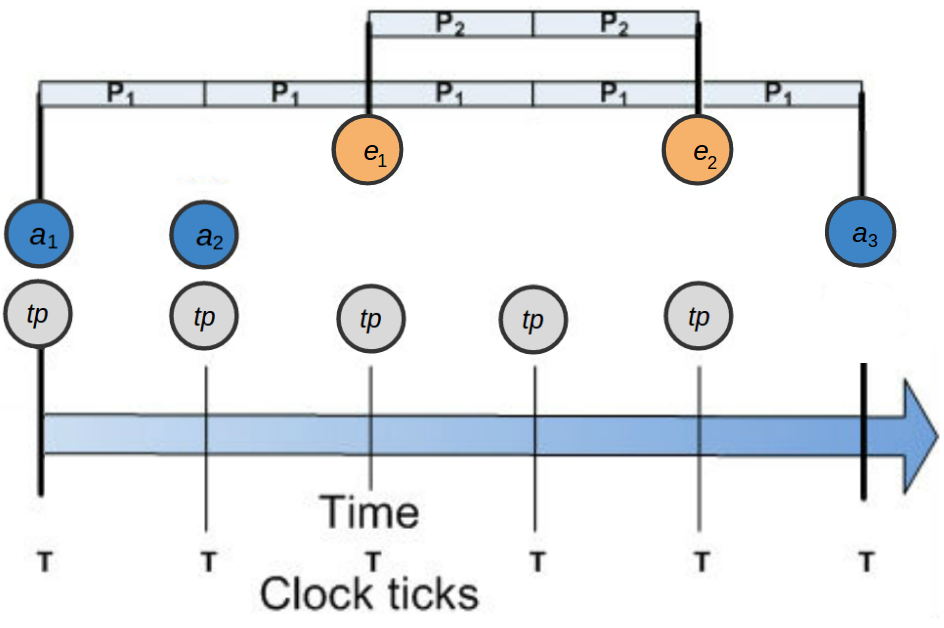}
  \caption{Uniform time discretization over a continuous model, showing timing and durations of happenings are directly tied to the time interval $\Delta t$. These happenings can only occur at specified timepoints $T$ separated by interval $\Delta t$ (i.e., actions $a_i$ and events $e_i$) or applied over the length of the time of the interval $\Delta t$ (i.e., processes $P_i$) using the special time passing action ($tp$).}
  \label{fig:discretized_timeline}
\end{figure}

\emph{Reconfiguration setup:} In reconfiguration, planning for scenarios 2 and 3, we consider an initial temporal horizon $T=1800s$, and a discretized time step $\Delta t=20s$. The temporal discretization dictates the accuracy of the dynamics and the state space size, thus this value needs to be fine enough to produce meaningful results while simultaneously ensuring solvability in a reasonable time. Figure~\ref{fig:discretized_timeline} outlines a continuous temporal timeline discretized using a uniform time interval. To demonstrate the impact of time discretization on performance and the resilience score, we also present results obtained using a time step $\Delta t=50s$ and $\Delta t=100s$. This is a crucial parameter in reconfiguration planning as it dictates the fault degradation update frequency and the frequency of decision points for reconfiguration actions\footnote{After time discretization, the reconfiguration actions can only be applied at multiples of the discretized time interval $\Delta t$. The system is assumed to be static over in between the decision points, and any updates due to events and processes are discrete.}. Overall, the more frequent the updates, the more accurate the discretized model of the system and its evolution over time, thus the resilience metric results are more reliable and trustworthy for finer time discretization. 

For a fair comparison, we consider identical planning parameters and constraints for both scenarios. We also use a resilience metric-based heuristic estimate $h(s)$ adjusted by the temporal information to bias the search towards states further into the plan (i.e. states for which more time has elapsed):
$$
h(s_t) = (\alpha_1 \mathcal{J}^1_t - \alpha_2 \mathcal{J}^2_t) \omega (T-t)/T
$$

\noindent $T$ is the temporal horizon, $s_t$ is a planning state at time $t$, $\mathcal{J}^1_t$ and $\mathcal{J}^2_t$ measure the deviation from the required massflow rates and fuel loss, respectively, over the horizon $t$. Weights $\alpha_1=0.5$ and $\alpha_2=0.5$ as in the original resilience metric equation, whereas $\omega=1.5$.

To further restrict the already vast search space we impose a constraint on the reconfiguration algorithm which limits the possible deviation from nominal fuel massflow rate at the engines to $\pm30\%$. For a fair comparison, we consider identical planning parameters and constraints for both scenarios. For the sake of runtime efficiency, here, we do not utilize the anytime algorithm, instead we only return the first encountered solution, regardless of its quality. For a fair comparison, we consider identical planning parameters and constraints for both scenarios.
\todo{Above P needs to be clarified}

In \emph{scenario 2}, without prognosis the reconfiguration module assumes a static leak intensity and no degradation over time. However, to maintain accuracy a new diagnosis is recomputed after each time step $\Delta t$ has elapsed. The reconfiguration planning process is repeated after each diagnosis with updated fault information (leak intensity) and a shorter time horizon $T' = T-\Delta t$. Actions performed during the first-elapsed time step from each reconfiguration iteration are appended to the final overall plan.
\emph{Scenario 3} does not require periodic re-planning as it incorporates the degradation model in the PDDL+ domain of the fuel system. However, re-planning would be required after an update of the fault degradation model.

Leak faults 1, 2, 7, and 8 are unrepairable by the reconfiguration module. This is due to the location of the leaks which, by design of the system, makes it impossible to isolate the fault and mitigate its detrimental effects. This can be communicated from the reconfiguration module when the search space is exhausted without finding a goal state. At that point, the vessel might request assistance in rectifying the fault from human operators. To avoid confusion, we denote such unrepairable cases as `X' when displaying the resilience results, whereas `-' signifies that there exists a plausible repair but the reconfiguration module was unable to find it in a reasonable amount of time.

\medskip

\emph{Scenario 1 results}: In the first scenario, no faults are tracked and no actions are taken to respond to the effects of these faults.
\begin{table}[ht!]
\centering
\resizebox{\columnwidth}{!}{%
\begin{tabular}{|c|c|c|c|c|c|c|c|c|c|}
\hline
\textbf{Degradation model} & \textbf{leak\_fault\_1} & \textbf{leak\_fault\_2} & \textbf{leak\_fault\_3} & \textbf{leak\_fault\_4} & \textbf{leak\_fault\_5} & \textbf{leak\_fault\_6} & \textbf{leak\_fault\_7} & \textbf{leak\_fault\_8} & \textbf{Average}\\
\hline
\textbf{Linear} & 0.65 & 0.65 & 0.65 & 0.65 & 0.64 & 0.64 & 0.66 & 0.66 & 0.649\\
\hline
\textbf{Exponential} & 0.66 & 0.66 & 0.66 & 0.66 & 0.66 & 0.66 & 0.68 & 0.68 & 0.676\\
\hline
\end{tabular}
}
\caption{Scenario 1: fuel system resilience results}
\label{tab:scenario_1_resilience}
\end{table}
The reduced resilience captures the loss of fuel at the leak points. As expected, in the case of the exponential degradation model, the resilience loss is more significant. In what follows, we show the impact of adapting the operation of the system to the detected leaks.

\begin{figure}[htp!]
  \centering
  \includegraphics[width=0.9\linewidth]{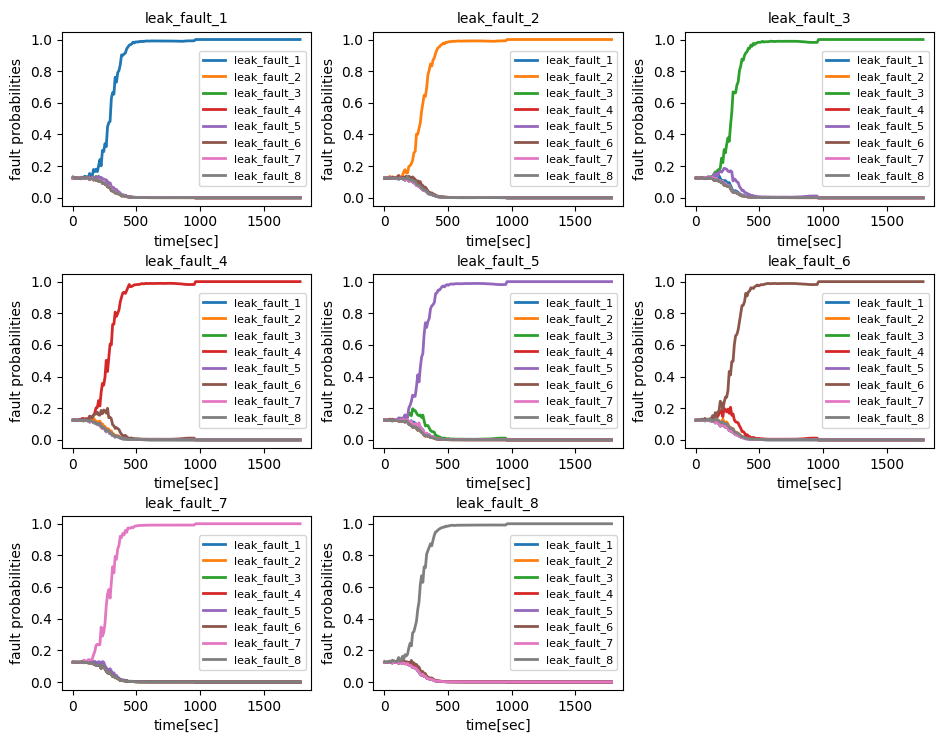}
  \caption{Diagnosis results: fault probabilities for the linear degradation model.}
  \label{fig:fault probabilities linear}
\end{figure}

\begin{figure}[htp!]
  \centering
  \includegraphics[width=0.9\linewidth]{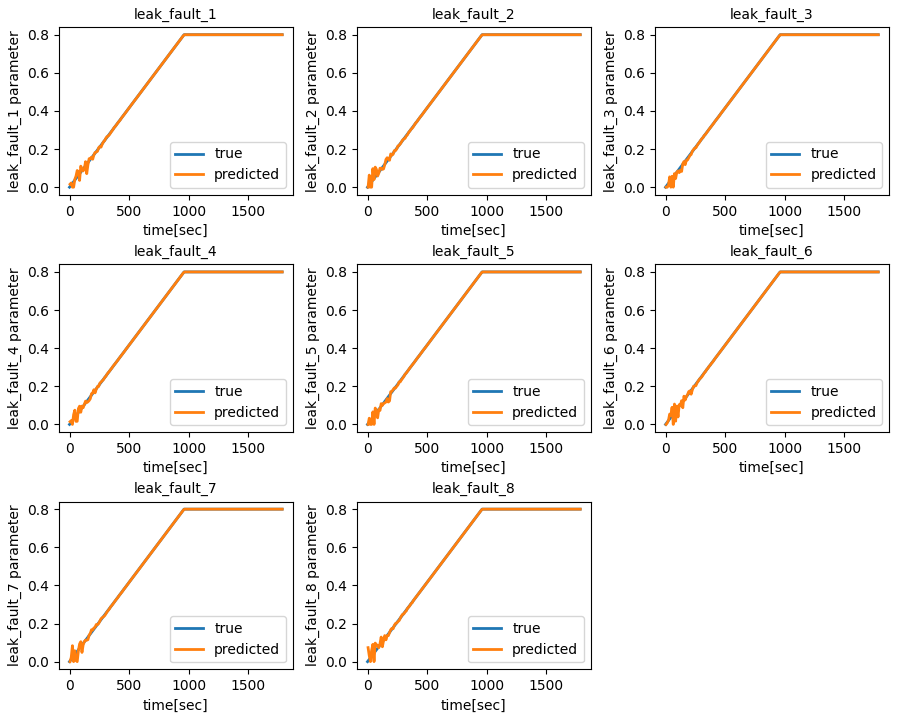}
  \caption{Diagnosis results: fault parameter estimates for the linear degradation model.}
  \label{fig:fault parameter linear}
\end{figure}

\begin{figure}[htp!]
  \centering
  \includegraphics[width=0.9\linewidth]{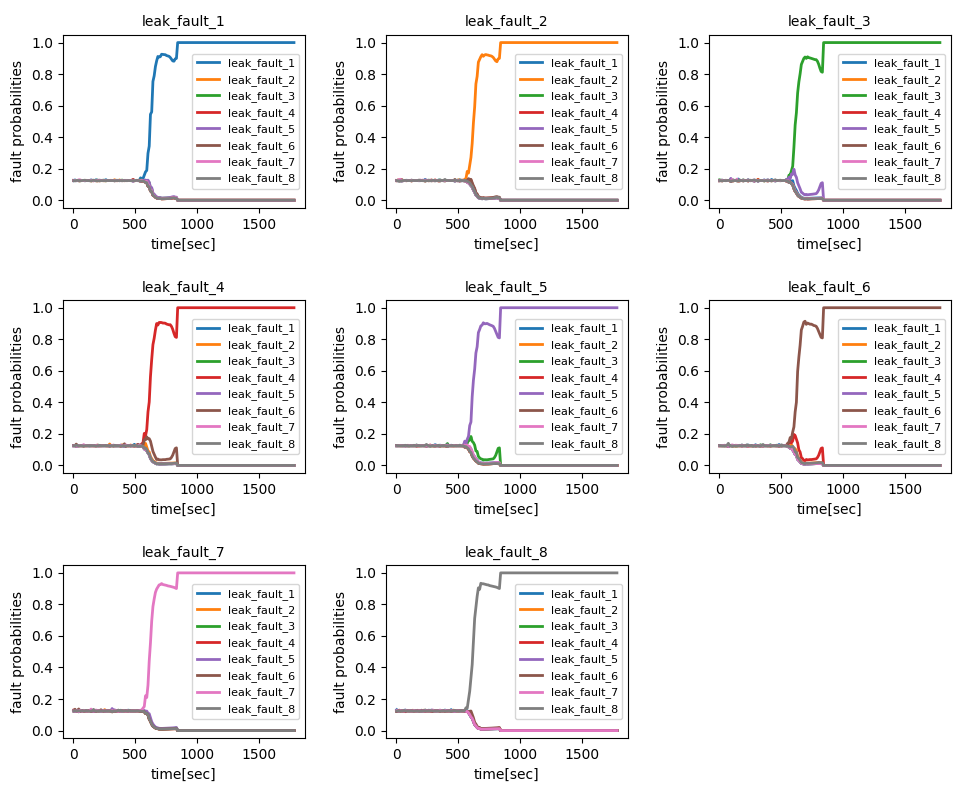}
  \caption{Diagnosis results: fault probabilities for the exponential degradation model.}
  \label{fig:fault probabilities exponential}
\end{figure}

\begin{figure}[htp!]
  \centering
  \includegraphics[width=0.9\linewidth]{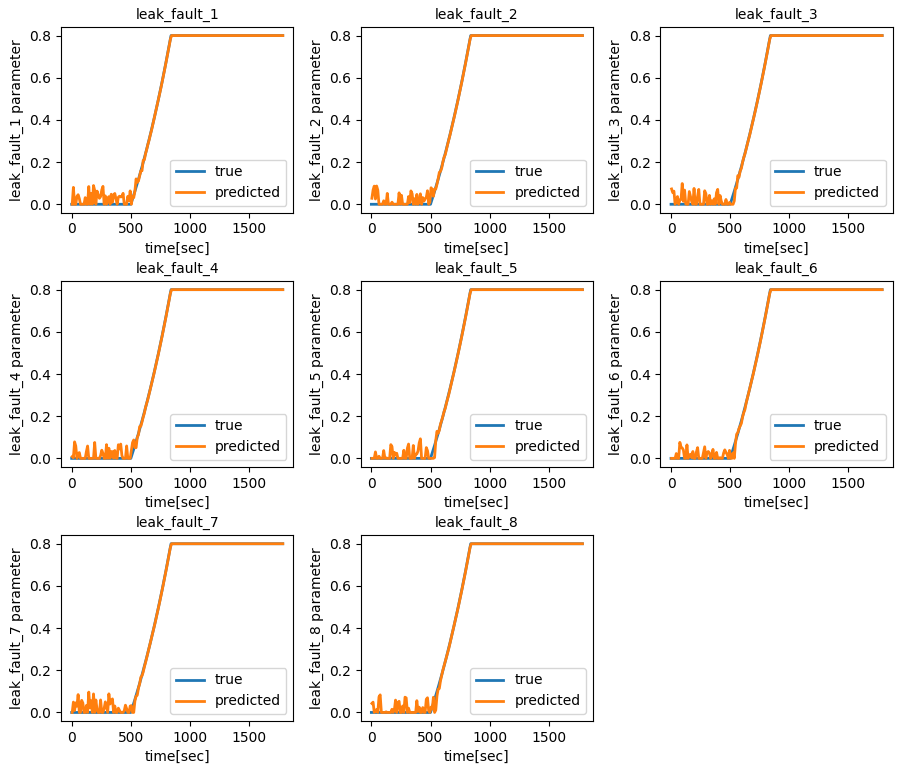}
  \caption{Diagnosis results: fault parameter estimates for the exponential degradation model.}
  \label{fig:fault parameter exponential}
\end{figure}

\emph{Scenario 2 results}: In this scenario, we assume we detect and isolate the leaks but use no prognosis. The reconfiguration algorithms assume that the fault magnitudes remain constant after each diagnosis. In other words, the reconfiguration algorithm does not make use of the predicted fault degradation models.
Figures \ref{fig:fault probabilities linear} and \ref{fig:fault probabilities exponential} depict the fault probabilities computed by the diagnosis algorithm, for the linear and exponential degradation models, respectively. For each degradation model, each of the eight figures corresponds to one of the eight single fault leak faults. In both cases, the fault probabilities of the true fault are dominant, once the fault magnitudes overcame the measurements noise. For each 180 windows, the diagnosis algorithm takes under 2 seconds to come up with a solution. Hence we are well under the 10 seconds windows size, indicating the feasibility of real-time implementation of the diagnosis module. Figures \ref{fig:fault parameter linear} and \ref{fig:fault parameter exponential} show the evolution of the fault parameter estimates. For small leaks, due to the measurement noise, the estimates oscillate close to zero in the beginning. As the magnitude of the leaks increase, the diagnosis algorithm is able to accurately track the true fault parameter, for both degradation models. Good fault estimates are imperative for training accurate prognosis models.

Reconfiguration in Scenario 2 using only diagnosis fails to improve the resilience of the system. The inability to solve the tasks using more accurate models can stem from the gradient of fault degradation, and the way the heuristic directs the search algorithm. In essence, the heuristic prioritizes states in which more time has elapsed. This, in turn, causes the reconfiguration actions to adjust the valve and pump settings to be applied later in the plan just before the system hits a critical point. However, without an accurate model of fault degradation the planner underestimates its evolution and postpones taking action until it is too late. After a re-diagnosis, it becomes clear that the fault has become uncontrollable and cannot be mitigated via reconfiguration. This is particularly visible when the fault degradation is updated more frequently at a smaller rate, with sparse steep increases in leak intensity, it is more likely that the planner will attempt to reconfigure earlier and make it before the point of no return.

The reconfiguration module only managed to solve two problems in each degradation fault model and only using the coarsest discretization. Thus the confidence in these results is low, since they lack accuracy. 

In addition to poor performance, diagnosis-only reconfiguration is also substantially more computationally inefficient. Since the reconfiguration module is unaware of the change in fault progression, each reconfiguration solution has to be re-computed for the remaining time horizon when a new diagnosis is provided (i.e., after each time step $\Delta t$, reconfiguration is repeated time horizon $T=T-\Delta t$).

\begin{table}[ht!]
\centering
\resizebox{\columnwidth}{!}{%
\begin{tabular}{|c|c|c|c|c|c|c|c|c|c|c|}
\hline
\textbf{Degradation model} & \textbf{$\Delta t$} & \textbf{leak\_fault\_1} & \textbf{leak\_fault\_2} & \textbf{leak\_fault\_3} & \textbf{leak\_fault\_4} & \textbf{leak\_fault\_5} & \textbf{leak\_fault\_6} & \textbf{leak\_fault\_7} & \textbf{leak\_fault\_8} & \textbf{Average}\\
\hline
\textbf{Linear} & 20s & 0.65$^\times$ & 0.65$^\times$ & 0.65* & 0.65* & 0.64* & 0.64* & 0.66$^\times$ & 0.66$^\times$ & 0.649\\
\hline
\textbf{Exponential} & 20s & 0.66$^\times$ & 0.66$^\times$& 0.66* & 0.66* & 0.66* & 0.66* & 0.68$^\times$ & 0.68$^\times$& 0.676\\
\hline
\textbf{Linear} & 50s & 0.65$^\times$ & 0.65$^\times$ & 0.65* & 0.65* & 0.64* & 0.64* & 0.66$^\times$ & 0.66$^\times$ & 0.649\\
\hline
\textbf{Exponential} & 50s & 0.66$^\times$ & 0.66$^\times$ & 0.66* & 0.66* & 0.66* & 0.66* & 0.68$^\times$ & 0.68$^\times$& 0.676\\
\hline
\textbf{Linear} & 100s & 0.65$^\times$ & 0.65$^\times$ & \cellcolor{green!25}\textbf{0.758} & \cellcolor{green!25}\textbf{0.758} & 0.64* & 0.64* & 0.66$^\times$ & 0.66$^\times$ & \cellcolor{green!25}\textbf{0.677}\\
\hline
\textbf{Exponential} & 100s & 0.66$^\times$ & 0.66$^\times$ & 0.66* & 0.66* & \cellcolor{green!25}\textbf{0.937} & \cellcolor{green!25}\textbf{0.937} & 0.68$^\times$ & 0.68$^\times$ & \cellcolor{green!25}\textbf{0.735}\\
\hline
\end{tabular}
}
\caption{Scenario 2: fuel system resilience results with reconfiguration based on diagnosis only. Green cell background indicates that the resilience score improved with reconfiguration, red cell indicates decrease in resilience with reconfiguration and white cell indicates no change (`$^\times$' denotes no reconfiguration possible, `*' denotes that the planner was unable to find a solution).}
\label{tab:scenario_2_resilience}
\end{table}

\begin{table}[ht!]
\centering
\resizebox{\columnwidth}{!}{%
\begin{tabular}{|c|c|c|c|c|c|c|c|c|c|c|}
\hline
\textbf{Degradation model} & \textbf{$\Delta t$} & \textbf{leak\_fault\_1} & \textbf{leak\_fault\_2} & \textbf{leak\_fault\_3} & \textbf{leak\_fault\_4} & \textbf{leak\_fault\_5} & \textbf{leak\_fault\_6} & \textbf{leak\_fault\_7} & \textbf{leak\_fault\_8} & \textbf{Average}\\
\hline
\textbf{Linear} & 20s & 0.65$^\times$ & 0.65$^\times$ & \cellcolor{green!25}\textbf{0.688} & \cellcolor{green!25}\textbf{0.662} & \cellcolor{green!25}\textbf{0.911} & \cellcolor{green!25}\textbf{0.911} & 0.66$^\times$ & 0.66$^\times$ & \cellcolor{green!25}\textbf{0.724}\\
\hline
\textbf{Exponential} & 20s & 0.66$^\times$ & 0.66$^\times$ & \cellcolor{red!25}\textbf{0.626} & \cellcolor{red!25}\textbf{0.626} & \cellcolor{green!25}\textbf{0.927} & \cellcolor{green!25}\textbf{0.927} & 0.68$^\times$ & 0.68$^\times$ & \cellcolor{green!25}\textbf{0.723}\\
\hline
\textbf{Linear} & 50s & 0.65$^\times$ & 0.65$^\times$ & \cellcolor{green!25}\textbf{0.721} & \cellcolor{green!25}\textbf{0.721} & \cellcolor{green!25}\textbf{0.915} & \cellcolor{green!25}\textbf{0.915} & 0.66$^\times$ & 0.66$^\times$ & \cellcolor{green!25}\textbf{0.736}\\
\hline
\textbf{Exponential} & 50s & 0.66$^\times$ & 0.66$^\times$ & \cellcolor{red!25}\textbf{0.652} & \cellcolor{red!25}\textbf{0.652} & \cellcolor{green!25}\textbf{0.930} & \cellcolor{green!25}\textbf{0.930} & 0.68$^\times$ & 0.68$^\times$ & \cellcolor{green!25}\textbf{0.731}\\
\hline
\textbf{Linear} & 100s & 0.65$^\times$ & 0.65$^\times$ & \cellcolor{green!25}\textbf{0.758} & \cellcolor{green!25}\textbf{0.758} & \cellcolor{green!25}\textbf{0.917} & \cellcolor{green!25}\textbf{0.917} & 0.66$^\times$ & 0.66$^\times$& \cellcolor{green!25}\textbf{0.746}\\
\hline
\textbf{Exponential} & 100s & 0.66$^\times$ & 0.66$^\times$ & \cellcolor{green!25}\textbf{0.717} & \cellcolor{green!25}\textbf{0.717} & \cellcolor{green!25}\textbf{0.937} & \cellcolor{green!25}\textbf{0.937} & 0.68$^\times$ & 0.68$^\times$ & \cellcolor{green!25}\textbf{0.748}\\
\hline
\end{tabular}
}
\caption{Scenario 3: fuel system resilience results with reconfiguration based on diagnosis and prognosis. Green cell background indicates that the resilience score improved with reconfiguration, red cell indicates decrease in resilience with reconfiguration and white cell indicates no change (`$^\times$' denotes no reconfiguration possible, `*' denotes that the planner was unable to find a solution).}
\label{tab:scenario_3_resilience}
\end{table}

\emph{Scenario 3 results}: In this scenario, the reconfiguration module has found a feasible plan to all solvable tasks proving that prognosis is a crucial piece of information that can significantly increase the resilience of a system. The resilience metric scores shown in Table \ref{tab:scenario_3_resilience} confirm that planning with coarser time discretization reduces the accuracy of the resilience scores since it underestimates the fault degradation rate (i.e., assumes a steady state condition for longer time intervals between updates). This can be seen with the resilience metric scores increasing in the opposite direction of the planning model granularity.

However, while the reconfiguration module ensures high resilience for leaks 5 and 6, leaks 3 and 4 score low. This is due to the fact that this leak cannot be fully isolated immediately as it would prevent the engines from being supplied fuel at an adequate massflow rate.  The planner must sacrifice some fuel through the leak in order to deliver enough fuel to the engines. The planner further compounds the issues by initially increasing the pump speed to compensate for the fuel leaking out, before isolating the leak and cutting off one of the tanks. As a result, this reconfiguration strategy has very little deviation from the nominal reference mass flow rates but suffers vast fuel loss which significantly reduces the resilience score. Such cases provide proof for more research into approaches to PDDL+ planning problems that focus on improving solution quality.

Using both diagnosis and prognosis significantly reduced the computational effort needed to find a suitable reconfiguration. Diagnosis and prognosis information is contained within a single reconfiguration planning problem, maintaining accuracy throughout the duration of the mission. In contrast with Scenario 2 (reconfiguration with diagnosis information only) no re-diagnosis and re-planning is required. Instead focus can be turned towards increasing reconfiguration planning accuracy by utilizing a finer time discretization. 

\begin{table}[ht!]
\centering
\label{tab:scenario_3_runtimes}
\begin{tabular}{|l|l|l|l|}
\hline
Fault deg & $\Delta t$ & Leak 3/4 & Leak 5/6 \\ \hline
Linear    & 20s        & 1080.3   & 97.1     \\ \hline
Nonlinear & 20s        & 103.8    & 214.0    \\ \hline
Linear    & 50s        & 114.6    & 56.75    \\ \hline
Nonlinear & 50s        & 31.9     & 73.5     \\ \hline
Linear    & 100s       & 7.3      & 221.9    \\ \hline
Nonlinear & 100s       & 28.6     & 167.2    \\ \hline
\end{tabular}%
\caption{Reconfiguration planning runtimes in seconds.}
\end{table}


\section{Conclusions}

This paper has shown that designing a resilient is system not
simply a matter of including a module for ``resilience''.  Rather,
resilience must be built in at the ground level for the system to be
able to be as successful as possible at achieving its mission in the
face of unforeseen faults and damage.  Each of the modules must be
adapted from what one sees in conventional design, and all these
adaptations must be carefully coordinated to achieve resilience.

Our run-time architecture requires (1) a diagnosis module that detects
and isolates any fault using sensor data, (2) a disambiguation module
that further isolates the fault by manipulating control inputs, (3) a
prognostics module that predicts fault progression, and (4) a
reconfiguration module which activates internal redundancy to ensure
mission success.  To achieve these we used parameter estimation,
regression and AI planning.  We could have used alternative approaches
to building these 4 modules, however, there is no avoiding needing the
modules.  We have shown that each of the modules is needed to achieve
the best system resilience.

We have shown that a model-based approach allows all designs use
exactly the same run-time and design-time software.  In our approach,
we only need to be provided a system description represented in a
systems modeling language (such as Modelica) and a PDDL description of
system goals, constraints and possible actions.  This description is
relatively simple because system dynamics is encoded through use of
the FMU (of the Modelica model).  From this model and PDDL description
provided at design time we fully automatically compile the data
structures required for run-time operation.  In particular, we build
an FMU, decompose the model and construct surrogate models for
efficiency, and build the PDDL structures needed for planning.

\bibliographystyle{alpha}
\bibliography{references}

\end{document}